\documentclass{article}

\PassOptionsToPackage{numbers, sort&compress}{natbib}

\usepackage[preprint]{neurips_2025}

\usepackage[utf8]{inputenc} 
\usepackage[T1]{fontenc}    

\usepackage{times}
\usepackage{epsfig}
\usepackage{algorithm}
\usepackage[noend]{algorithmic}
\usepackage{graphicx}
\renewcommand{\algorithmiccomment}[1]{\bgroup\hfill $\triangleright$ ~#1\egroup}
\usepackage{amsmath}
\usepackage{amssymb}
\usepackage{amsfonts}       
\usepackage{amsthm}         
\usepackage{nicefrac}       
\usepackage{bm}
\usepackage{mathtools}

\usepackage{pgfplots}
\usepackage{pgfplotstable}
\pgfplotsset{compat=newest}
\usepackage{color}
\usepackage{xcolor}
\usepackage{booktabs,siunitx} 
\usepackage{makecell}
\usepackage{enumitem}
\usepackage[percent]{overpic}
\usepackage[caption=false]{subfig}
\usepackage{rotating}
\usepackage[misc]{ifsym}
\usepackage{multirow}

\usepackage[pagebackref,breaklinks=true,colorlinks,bookmarks=false]{hyperref} 
\usepackage{url}            
\usepackage{microtype}      

\usepackage{amsmath,amsfonts,bm}

\def\eqref#1{(\ref{#1})}

\def\1{\bm{1}}

\DeclareMathAlphabet{\mathsfit}{\encodingdefault}{\sfdefault}{m}{sl}
\SetMathAlphabet{\mathsfit}{bold}{\encodingdefault}{\sfdefault}{bx}{n}

\newcommand{\best}[1]{\textbf{#1}}
\newcommand{\scnd}[1]{\underline{#1}}

\newcommand{\PAR}[1]{\noindent{\bf #1~}}

\definecolor{grid-search}                   {RGB}{227,111, 38}
\definecolor{yoto-negentsoftplus}           {RGB}{ 62,135,207}

\newcommand{\Ours}{YOTO}

\title{You Only Train Once}

\author{%
  Christos Sakaridis\\
  ETH Z\"urich\\
  \texttt{csakarid@ethz.ch}}

\begin{document}

\maketitle

\begin{abstract}
The title of this paper is perhaps an overclaim. Of course, the process of creating and optimizing a learned model inevitably involves multiple training runs which potentially feature different architectural designs, input and output encodings, and losses. However, our method, You Only Train Once (\Ours{}), indeed contributes to limiting training to one shot for the latter aspect of losses selection and weighting. We achieve this by automatically optimizing loss weight hyperparameters of learned models in one shot via standard gradient-based optimization, treating these hyperparameters as regular parameters of the networks and learning them. To this end, we leverage the differentiability of the composite loss formulation which is widely used for optimizing multiple empirical losses simultaneously and model it as a novel layer which is parameterized with a softmax operation that satisfies the inherent positivity constraints on loss hyperparameters while avoiding degenerate empirical gradients. We complete our joint end-to-end optimization scheme by defining a novel regularization loss on the learned hyperparameters, which models a uniformity prior among the employed losses while ensuring boundedness of the identified optima. We evidence the efficacy of \Ours{} in jointly optimizing loss hyperparameters and regular model parameters in one shot by comparing it to the commonly used brute-force grid search across state-of-the-art networks solving two key problems in computer vision, i.e.\ 3D estimation and semantic segmentation, and showing that it consistently outperforms the best grid-search model on unseen test data. Code will be made publicly available.
\end{abstract}

\section{Introduction}
\label{sec:intro}

Artificial intelligence has been revived in recent years via very large hierarchical parametric models in the form of neural networks, whose success is largely due to the invention of effective and efficient algorithms for optimizing them~\cite{rumelhart1986learning}. The most widely used class of such algorithms is based on gradient descent (GD), because gradients of the minimization objective with respect to model parameters can be computed fast and thus afford numerous rapid iterations of the algorithm, which helps to quickly improve the objective even with stochastic updates that use very small fractions of the dataset~\cite{bottou1998online,lecun1998efficient} and to converge to fair local optima.

While optimizing a given instance of a neural network architecture on a given data-based objective, i.e.\ a \emph{given empirical loss}, is thus by now well-understood and solvable, (i) selecting an optimal architecture itself as well as (ii) optimally \emph{selecting and weighting empirical losses themselves still poses fundamental challenges} to the community. The first of these two points has given rise to the relatively new research area of neural architecture search (NAS)~\cite{zoph2017neural,liu2018darts}. Moreover, both (i) and (ii) involve the optimization of \emph{hyperparameters} (HPs), i.e.\ parameters which are of a different, potentially non-differentiable or even non-continuous nature than regular parameters of neural network layers. Hyperparameter optimization (HPO)~\cite{maclaurin2015gradient,pedregosa2016hyperparameter,bengio2000gradient,Rajeswaran2019implicitgrad} consists precisely in optimizing such higher-level parameters of either the architecture or the losses. However, the crux of both NAS and HPO in practice is that they typically require multiple complete training runs in order to first compute the performance of the network at inference before making a single iteration in the architecture space or the HP space. Each outer iteration of this meta-optimization is thus extremely inefficient compared to inner, gradient-based iterations in the parameter space.

In this work, we focus on point (ii), i.e.\ on the selection and weighting of losses, starting with the observation that a large portion of HPO experiments in practice lies in the optimization of \emph{HPs which weigh empirical losses} in the common case of the presence of multiple distinct such objectives and typically appear as weights of a linear combination of losses. A piece of evidence for the ubiquity of this setting comes from randomly examining the 20 papers which appear at the top of the online list of the CVPR~2024 proceedings~\cite{cvpr2024proceedings}: 10 out of the 20 optimize such a linear combination of distinct empirical losses. More broadly, notable vision models that fall in this setting include Gaussian Splatting~\cite{kerbl2023gs}, pix2pix~\cite{isola2017image}, the R-CNN series~\cite{fast:rcnn,faster:rcnn,he2017maskrcnn}, and DETR~\cite{carion2020detr}, and respective language models include PaLM~\cite{chowdhery2023palm}, Switch Transformers~\cite{fedus2022switch}, and DistilBERT~\cite{sanh2019distilbert}. Because of the moderate dimensionality of this HP space and the aforementioned typical inefficiency and intensity of sophisticated HPO techniques, practitioners in learning areas such as vision or language commonly resort to simple yet computationally intensive brute-force approaches, such as grid search in this space~\cite{fedus2022switch}, to optimize these loss HPs. However, such approaches suffer from the curse of dimensionality, which makes them cumbersome even for optimizing two HPs when a suitable range of values is not known a priori. Can we instead directly optimize loss weight HPs simultaneously and jointly with regular model parameters \emph{in one shot} building on standard gradient-based methods?

We present a novel optimization algorithm, named You Only Train Once (\Ours{}), and experimental validation of it to answer positively to this question. \Ours{} hinges on the linearity of the above composite empirical loss to the involved loss weight HPs, which allows us to express this loss as the ultimate, differentiable layer of the overall end-to-end model. In turn, gradients with respect to the loss HPs can be computed and backpropagated in order to update the latter together with the regular parameters of the network, i.e.\ to apply standard gradient-based learning on them. We bake the inherent positivity of loss weights in our novel loss layer by operating in a logarithmic space, and decouple the loss scale from the learning rate via a softmax parameterization which ensures normalization. Akin to weight decay~\cite{loshchilov2017adamw} used for regular network weights which reside in a Euclidean space, we complete \Ours{} with a novel hyperparameter decay for regularization, which uses the gradient of a negated entropy term and a softplus term that promote uniformity and upper-boundedness, respectively. We surprisingly evidence through experiments on the key learning field of vision using two central tasks how \emph{letting the loss HPs of neural networks be jointly optimized with the regular model parameters} via normal gradient-based optimization delivers overall models that not only match but \emph{exceed the generalization capability of gradient-based optimization applied only to the regular parameters}, in which loss HPs are ``segregated'' from the former weights, kept fixed in each optimization run, and optimized by brute-force or meta-learning-based approaches that are still computationally far more intensive and slow. Last but not least, \Ours{} exhibits fair robustness to stochasticity and initialization.

\section{Background and Related Work}
\label{sec:related}

HPO is as old as machine learning itself~\cite{akaike1974new,box1951experimental}, as it arises from the ubiquitous empirical need for an optimal experimental design and thus plays a critical role in the field. HPs not only determine the generalization capabilities of trained models, but may actually decide which method constitutes the state of the art. The distinction between standard optimization of parametric models and HPO has been invariably based on the \emph{nested} nature of the latter~\cite{franceschi2024hyperparameter}. Formally, let $f(\boldsymbol{\lambda})$ denote a learned mapping configured by the HP vector $\boldsymbol{\lambda}$ and consider a function $\mathcal{M}$ that evaluates the performance of $f$ on a learning task. The composition $h(\boldsymbol{\lambda}) = \mathcal{M} \circ f(\boldsymbol{\lambda})$ is referred to as the response function. In this setting, HPO is invariably viewed as optimizing $h$ with respect to $\boldsymbol{\lambda}$ in the literature~\cite{bengio2000gradient,snoek2012practical,maclaurin2015gradient,klein2017fast,lorenzo2017particle}. Under this regime, a diverse set of methods have been proposed for HPO, ranging from (i) gradient-based methods which compute approximate ``hypergradients'' of $h$ w.r.t. $\boldsymbol{\lambda}$~\cite{bengio2000gradient,pedregosa2016hyperparameter,larsen1998adaptive,franceschi2017forward,maclaurin2015gradient,franceschi18bilevel,grazzi2020iteration,franceschi2024hyperparameter,liao2018reviving}, to (ii) model-based methods~\cite{bergstra2011algorithms,hutter2011sequential,snoek2012practical,klein2017fast,falkner2018bohb}, which are intrinsically characterized by a sequential operation in exploring the HP space, and to (iii) population-based methods~\cite{hansen1996adapting,jaderberg2017population,lorenzo2017particle,loshchilov2016cma,tao2020learning}, which are by contrast parallel but require the computationally intensive maintenance of a multitude of concurrent HP estimates. Across all these classes of HPO as well as HPO methods that combine them~\cite{tao2020learning,falkner2018bohb}, one needs to run an \emph{entire optimization} of the mapping $f$ w.r.t.\ its regular parameters $\mathbf{w}$ on a standard empirical loss to fit $f$ to the available data, e.g.\ a full training run for a neural network, in order to evaluate the response function $h$ at a \emph{single} point $\boldsymbol{\lambda}$ in the HP space and optimize the outer objective, in what is known as \emph{bi-level optimization}. Multi-fidelity HPO~\cite{li2018hyperband,li2018massively,karnin2013almost,jamieson2016nonstochastic,klein2017fast,falkner2018bohb,swersky2013multi,swersky2014freeze,bornschein2020small} attempts to mitigate the fundamental inefficiency of bi-level optimization by tuning the regular parameters $\mathbf{w}$ in the main optimization faster via early stopping or training on partial data, but does not cancel the fact that a single update in the HP space is orders of magnitude slower than an update of regular parameters, e.g.\ via GD.

The standard argument for justifying the disjoint, bi-level optimization of regular parameters and HPs is that the analytical form of $h$ is typically either unknown or impractical to handle. Yet these problems also exist for the regular parameters $\mathbf{w}$ which the learned mapping $f$ needs to optimize for and they have been largely solved for complex differentiable models by backpropagation-based GD. While the training objectives of mappings $f$ cannot be differentiated with respect to all their HPs, we recognize that this possibility does exist for the weights $\boldsymbol{\lambda} \in \mathbb{R}^{K+1}$ of the widely used composite empirical loss in \eqref{eq:empirical:loss:composite} below. Moreover, the regular parameters $\mathbf{w}$ are typically also not tuned directly on the performance function $\mathcal{M}$, but rather on the original optimization objective or loss. Thus, there is no fundamental reason preventing us from rethinking HPO, treating these loss weight HPs $\boldsymbol{\lambda}$ as regular parameters of $f$ and optimizing them regularly on the same empirical loss as regular parameters rather than on $\mathcal{M}$.

Particularly related to our optimization of a composite loss function with multiple independent components is multi-objective HPO~\cite{emmerich2011hypervolume,keane2006statistical,knowles2006parego,hernandez2016predictive,zhang2020random,belakaria2019max}. A key difference is that these works focus rather on multiple \emph{performance} or validation-level objectives than on multiple \emph{training-level} objectives, i.e.\ empirical losses. Also related is constrained HPO~\cite{gelbart2014bayesian,gardner2014bayesian,letham2019bayesian,perrone2019constrained}, which typically expresses constraints in terms of the value of the validation-level objective and thereby indirectly constrains the feasible set of HPs, whereas our optimized loss HPs have direct intrinsic constraints on their values per se, as they need to be positive. Finally, our work is related to NAS~\cite{zoph2017neural,yang2020nasefh,Zhou2020econas,dong2021autohas,izquierdo2021bagofbaselines,Zela2018TowardsAuto,Dong2019searching,Dai2021FBNetv3,Ren2021NASsurvey,Elsken2019NASsurvey} in the wide sense and shares analogies with methods from that area proposing a differentiable formulation~\cite{liu2018darts,Zela2020Understanding}, but we solely focus on continuous HPs involved in the optimized loss rather than on discrete or categorical parameters involved in the models' modules.

\section{Loss Hyperparameter Optimization via Standard Gradient-Based Learning}
\label{sec:method}

\subsection{Preliminaries}
\label{sec:method:preliminaries}

In the design of parametric learned mappings $f$ which are optimized by minimizing a basic empirical loss $l_0$ with respect to the parameters $\mathbf{w}$ of $f$, practitioners often introduce additional loss terms in the overall empirical loss $l_i,\,i \in \{1,\dots,K\}$, which typically play an auxiliary role and help obtain mappings that generalize better to unseen data during inference. The overall empirical loss $L_e$ becomes
\begin{equation} \label{eq:empirical:loss:composite}
    L_e(f(\mathbf{w})|\boldsymbol{\lambda}) = \sum_{i=0}^K \lambda_i l_i(f(\mathbf{w})),
\end{equation}
where $\lambda_i > 0,\,i \in \{0,\dots,K\}$, are positive HPs which serve as weights of the respective loss terms and which also need to be optimized besides the regular parameters $\mathbf{w}$ of the mapping. We term the empirical loss $L_e$ of \eqref{eq:empirical:loss:composite} as \emph{composite empirical loss}. However, the standard practice is to optimize $\boldsymbol{\lambda}$ on the \emph{validation} set, based on the final performance metrics of $f(\mathbf{w})$ having been optimized on the \emph{training} set, e.g.\ through GD. The former optimization of $\boldsymbol{\lambda}$ typically involves a non-end-to-end, inefficient search over several fixed values of $\boldsymbol{\lambda}$, which suffers from the curse of dimensionality w.r.t.\ the number $K+1$ of HPs.

\subsection{Composite Loss Layer}
\label{sec:method:composite:loss:layer}

First, we recognize that the common formulation of the composite empirical loss in \eqref{eq:empirical:loss:composite} is essentially a linear parametric mapping of the $K+1$ inputs $l_i(f(\mathbf{w}))$ to the output $L_e$, parameterized by $\boldsymbol{\lambda}$. Thus, we can \emph{extend} the typical end-to-end formulation of parametric mappings from the commonly defined end of predictions $f(\mathbf{w})$ to the ultimate end of $L_e$ itself as
\begin{equation} \label{eq:empirical:loss:composite:layer}
    L_e = g(\boldsymbol{\lambda}) \circ{} \mathbf{l} \circ{} f(\mathbf{w}) = g(\mathbf{l}(f(\mathbf{w})),\boldsymbol{\lambda}) = \boldsymbol{\lambda}^T\mathbf{l}(f(\mathbf{w})),
\end{equation}
where $\mathbf{l}(f(\mathbf{w})) = \left(l_0(f(\mathbf{w})),\,\dots,\,l_K(f(\mathbf{w})\right)^T$. We term $g$ the composite loss layer. Since $g$ is linear, it is \emph{differentiable}, hence amenable to standard gradient-descent-based methods and backpropagation for optimization. That is, we can optimize the HPs $\boldsymbol{\lambda}$ of $g$ as regular parameters at each iteration of gradient-based optimization, together with the regular parameters $\mathbf{w}$ of $f$. This paradigm shift leaves the computation of the gradients of the regular parameters $\mathbf{w}$ w.r.t.\ $L_e$ unchanged, as
\begin{equation} \label{eq:w:grad:empirical}
    \frac{\partial{}L_e}{\partial{}\mathbf{w}} = \frac{\partial{}\left(\boldsymbol{\lambda}^T\mathbf{l}(f(\mathbf{w}))\right)}{\partial{}\mathbf{w}} = \boldsymbol{\lambda}^T\frac{\partial{}\mathbf{l}(f(\mathbf{w}))}{\partial{}\mathbf{w}} = \sum_{i=0}^K \lambda_i \frac{\partial{}l_i(f(\mathbf{w}))}{\partial{}\mathbf{w}}.
\end{equation}
However, because of the positivity constraints $\lambda_i > 0,\,i \in \{0,\dots,K\}$, we cannot directly apply standard unconstrained gradient optimization on $\boldsymbol{\lambda} \in \mathbb{R}^{K+1}$ merely with the formulation of \eqref{eq:empirical:loss:composite:layer}.

\subsection{Softmax Parameterization}
\label{sec:method:softmax}

Instead of directly learning the weights $\lambda_i$ themselves, one can learn the exponents
\begin{equation} \label{eq:mus}
    \mu_i = \log(\lambda_i) \in \mathbb{R},\,i \in \{0,\,\dots,\,K\},
\end{equation}
which are amenable to unconstrained gradient-based optimization. This ensures that the weights, which now become exponentials $\lambda_i = \exp(\mu_i)$, are positive by definition.

Nonetheless, this exponential trick alone makes the gradients w.r.t.\ the HPs degenerate, in particular positive. More formally, if we substitute \eqref{eq:mus} into \eqref{eq:empirical:loss:composite} and differentiate w.r.t.\ $\mu_i$, we obtain $\frac{\partial{}L_e}{\partial{}\mu_i} = \exp(\mu_i)l_i(f(\mathbf{w})) > 0$, which implies that all gradient-based updates will reduce the exponents $\mu_i$ and hence the weights $\lambda_i$. Intuitively, there is the incentive to reduce the weight of each term of the composite loss, so that the overall loss $L_e$ is reduced too.

We circumvent this degeneracy by introducing a softmax parameterization of the composite loss layer. This parameterization adopts the exponential trick above, but also additionally introduces a normalization factor, which creates competition between different HPs and loss terms. More formally, we formulate the composite loss layer as
\begin{equation} \label{eq:softmax}
    L_e = \left. \sum_{i=0}^K \exp(\mu_i)l_i(f(\mathbf{w})) \middle/ \displaystyle\sum_{j=0}^K \exp(\mu_j) \right. = \left(\text{Softmax}(\boldsymbol{\mu})\right)^T\mathbf{l}(f(\mathbf{w})).
\end{equation}
For the weights $\boldsymbol{\lambda} = \text{Softmax}(\boldsymbol{\mu})$, besides the preservation of the positivity constraints, we also have $\sum_{i=0}^K \lambda_i = 1$, implying that the composite loss $L_e$ now becomes a \emph{convex combination} of the individual loss terms $l_i$.
With the softmax parameterization of \eqref{eq:softmax}, the HP gradients become
\begin{equation} \label{eq:mu:grad:empirical}
    \frac{\partial{}L_e}{\partial{}\mu_i} = \left. \exp(\mu_i)\left(\displaystyle\sum_{\substack{j=0 \\ j \neq i}}^K\left(l_i(f(\mathbf{w})) - l_j(f(\mathbf{w}))\right)\exp(\mu_j)\right) \middle/ \left(\displaystyle\sum_{j=0}^K \exp(\mu_j)\right)^2 \right. .
\end{equation}
The proof of \eqref{eq:mu:grad:empirical} is in Appendix~\ref{sec:appendix:derivation:empirical}. Because of the differences $l_i(f(\mathbf{w})) - l_j(f(\mathbf{w}))$ which appear in the numerator on the RHS of \eqref{eq:mu:grad:empirical}, the HP gradients can be either positive or negative, thus solving the above degeneracy. As what is important to learn in the formulation of \eqref{eq:softmax} for the composite loss layer are the \emph{relative} values of the exponents $\mu_i$, the associated degrees of freedom are actually $K$. Thus, we freeze the exponent $\mu_0$ corresponding to the basic loss $l_0$ to $0$ throughout the optimization and only learn $\mu_i$ for $i \in \{1,\,\dots,\,K\}$, i.e.\ for the auxiliary losses, using the gradients from \eqref{eq:mu:grad:empirical}.

Another important attribute of the softmax parameterization is that it preserves the \emph{scale} of the composite loss to the same level as that of the basic loss $l_0$. The basic loss $l_0$ is typically the starting point in the design of a mapping which may subsequently include auxiliary losses $l_i, i \geq 1$. The scale of $l_0$ affects the optimal scale of the learning rate $\eta$, because (i) the gradient is a linear operator and directly inherits a change of scale in the loss, and (ii) the learning rate acts as a multiplier of the loss gradient, so a change of scale in the loss effectively acts as a change of scale in the learning rate. By merely modifying $l_0$ to a composite loss $L_e$ via \eqref{eq:empirical:loss:composite} without constraining $\lambda_i$, one would change the scale of the overall loss by a factor of $\sum_{i=0}^K \lambda_i \neq 1$, effectively changing the learning rate. By contrast, our softmax parameterization preserves the original scale of the learning rate thanks to its convex combination character. This point is important in conducting comparisons including our approach on existing methods which originally do not apply this normalization and alter the effective learning rate when introducing auxiliary losses by setting $\sum_{i=0}^K \lambda_i \neq 1$. Being able to also automatically optimize the learning rate is an exciting related possibility, which we do not address here but leave for future work.

\subsection{Regularization}
\label{sec:method:regularization}

Our treatment of loss HPs as regular parameters optimized based on standard gradient updates necessitates the application of regularization to these HPs, similarly to what is standard practice for regular parameters. However, the difference in the mathematical functionality of our HP exponents $\boldsymbol{\mu}$ and of regular parameters $\mathbf{w}$ of the mapping which is learned, which typically reside in a standard Euclidean space, dictates a differentiation in their regularization. In the latter case, weights in $\mathbf{w}$ typically multiply features that assume arbitrary real values, so a regularity prior consists in small magnitudes and is typically pursued via an squared $L_2$ penalty which is equivalent to a \emph{weight decay} towards $\mathbf{0}$~\cite{loshchilov2017adamw}.

By contrast, our exponents $\mu_i$ are passed through a softmax to multiply positive loss values. Thus, the resulting softmax vector can be viewed as a discrete probability distribution over the various loss terms $l_i$, the data-related bias of which towards any individual loss can be regularized via the \emph{negated entropy} of the distribution that favors ``simpler'' distributions close to uniform. Moreover, to regularize the \emph{absolute} scale of the exponents $\mu_i$, to which negated entropy is invariant, we include a \emph{softplus} term for each learnable $\mu_i$ to our regularization loss $L_r$, so that the latter becomes
\begin{equation} \label{eq:loss:reg}
    L_r(\boldsymbol{\mu}) =  \rho\left(\sum_{i=0}^K \frac{\exp(\mu_i)}{\sum_{j=0}^K \exp(\mu_j)} \log\left(\frac{\exp(\mu_i)}{\sum_{j=0}^K \exp(\mu_j)}\right) + \sum_{i=1}^K \log\left(1 + \exp(\mu_i)\right)\right),
\end{equation}
where $\rho > 0$ is the single non-learnable \emph{hyperparameter decay} which our method introduces. In particular, $\rho$ is shared by all loss HPs, akin to the standard \emph{weight decay} $\lambda$~\cite{loshchilov2017adamw} which serves as a single non-learnable HP shared by all weights $\mathbf{w}$ of the model. In this way, we substitute the selection of multiple loss-specific HPs $\mu_i$ with a single generic HP decay $\rho$ having a range of values that is common across multiple diverse learned models and needing minimal tuning, similar to the standard treatment of weight decay of regular parameters in neural networks. Moreover, we keep the same weight decay or other regularization for $\mathbf{w}$ as the original optimization algorithm on which \Ours{} is implemented in each case, as \eqref{eq:loss:reg} only pertains to $\boldsymbol{\mu}$.

For the gradients of our regularizer w.r.t.\ the HPs, $\partial{}L_r/\partial{}\mu_i$, we prove in Appendix~\ref{sec:appendix:derivation:reg} that
\begin{equation} \label{eq:mu:grad:reg}
    \frac{1}{\rho}\frac{\partial{}L_r}{\partial{}\mu_i} = \frac{\exp(\mu_i)\left(\sum_{j=0}^K \exp(\mu_j)(\mu_i-\mu_j)\right)}{\left(\sum_{j=0}^K \exp(\mu_j)\right)^2} + \frac{\exp(\mu_i)}{1 + \exp(\mu_i)},\,i \in \{1,\,\dots,\,K\}.
\end{equation}

\begin{algorithm}[tb]
\caption{\Ours{} with SGDW with momentum}
\footnotesize
\label{alg:yoto:sgdw}
\begin{algorithmic}[1]
\STATE{\textbf{given} initial learning rate $\alpha \in \mathbb{R}$, momentum factor $\beta_1 \in \mathbb{R}$, weight decay $\lambda > 0$, hyperparameter decay $\rho > 0$, initialization parameter $0 < \epsilon \ll 1$}\label{alg:yoto:sgdw:given}
\STATE{\textbf{initialize} time step $t \leftarrow 0$, parameter vector $\mathbf{w}_{t=0} \in \mathbb{R}^n$, loss HP exponent vector $\boldsymbol{\mu}_{t=0} \leftarrow (0,\,\log(\epsilon),\,\dots,\,\log(\epsilon)) \in \mathbb{R}^{K+1}$, first moment vector for parameters $\mathbf{m}_{t=0} \leftarrow \mathbf{0}$, first moment vector for HPs $\mathbf{n}_{t=0} \leftarrow \mathbf{0}$, schedule multiplier $\eta_{t=0} \in \mathbb{R}$}
\REPEAT
	\STATE{$t \leftarrow t + 1$}
	\STATE{$\nabla{}L_{e,t}(\mathbf{w}_{t-1}),\nabla{}L_{e,t}(\boldsymbol{\mu}_{t-1}) \leftarrow  \text{SelectBatch}(\mathbf{w}_{t-1},\boldsymbol{\mu}_{t-1})$}  \COMMENT{compute empirical gradient, using i.a.\ \eqref{eq:mu:grad:empirical}}
    \STATE{$\mathbf{g}_t \leftarrow \nabla{}L_{e,t}(\mathbf{w}_{t-1})$}
    \STATE{$\mathbf{h}_t \leftarrow \nabla{}L_{e,t}(\boldsymbol{\mu}_{t-1})$}
	\STATE{$\eta_t \leftarrow \text{SetScheduleMultiplier}(t)$}
	\STATE{$\mathbf{m}_t \leftarrow \beta_1 \mathbf{m}_{t-1} + \eta_t \alpha \mathbf{g}_t $}
    \STATE{$\mathbf{n}_t \leftarrow \beta_1 \mathbf{n}_{t-1} + \eta_t \alpha \mathbf{h}_t $}
    \STATE{$\mathbf{w}_t \leftarrow \mathbf{w}_{t-1} - \mathbf{m}_t - \eta_t\alpha\lambda\mathbf{w}_{t-1}$}
    \STATE{$\boldsymbol{\mu}_t \leftarrow \boldsymbol{\mu}_{t-1} - \mathbf{n}_t - \eta_t\alpha\rho\nabla{}L_{r,t}(\boldsymbol{\mu}_{t-1})$  \COMMENT{include gradient of regularizer in update using \eqref{eq:mu:grad:reg}}}
\UNTIL{ \textit{stopping criterion is met} }
\RETURN{optimized parameters $\mathbf{w}_t$, optimized loss HPs $\boldsymbol{\mu}_t$}
\end{algorithmic}
\end{algorithm}

\subsection{Initialization and Overall \Ours{} Algorithm}
\label{sec:method:initialization}

While Sec.~\ref{sec:method:softmax} and \ref{sec:method:regularization} detail the gradients for updating loss HPs within a given iteration of gradient-based optimization, how to initialize the exponents $\mu_i,\,i \in \{1,\,\dots,\,K\}$, is also of central importance. The strategy that we follow in practice in our experiments is to initialize these $\mu_i$, which correspond to auxiliary empirical loss terms, uniformly with the same value $\mu_{i,0} \leftarrow \log(\epsilon)$, where $\epsilon \ll 1$ is a small positive number. Combined with the fact that $\mu_{0,t} = 0$ for the basic empirical loss $l_0$, this implies that we start optimization with weak overall contributions $\lambda_{i,0}l_i$ for the auxiliary empirical losses, i.e.\ for $i \in \{1,\,\dots,\,K\}$, and let the model potentially increase these contributions automatically in the case where such an increase is also beneficial for optimizing the more heavily weighted basic loss contribution, $\lambda_0 l_0$, or keep the former contributions at a low value throughout the optimization. In both cases, the model \emph{selects automatically} which losses are useful for optimization and to what degree. As we evidence in Sec.~\ref{sec:exp}, the optimal $\boldsymbol{\mu}$ and $\boldsymbol{\lambda}$ values after convergence of a certain network are rather stable across a wide range of initializations, hinting on the insensitivity of \Ours{} to the particular initialization. A full instance of our \Ours{} algorithm combined with SGD with decoupled decay (SGDW)~\cite{loshchilov2017adamw} is presented in Algorithm~\ref{alg:yoto:sgdw}.

\section{Experiments}
\label{sec:exp}

From the wide range of areas using models learned with gradient-based loss optimization, which are all relevant to \Ours{}, we select the key area of computer vision for experimentation with and validation of our method, in particular two of its central tasks in 3D and segmentation, i.e.\ monocular 3D/depth estimation and semantic segmentation. The total used compute is $\approx$1,150 RTX4090 days.

\subsection{3D Estimation}
\label{sec:exp:3d}

We first apply \Ours{} to the state-of-the-art monocular metric 3D/depth estimation method of UniDepth~\cite{piccinelli2024unidepth} for optimizing its loss HPs. In particular, UniDepth originally optimizes a composite empirical loss $L_e$ based on its metric dense 3D predictions $(\mathbf{Z}_{\text{log}},\mathbf{\Theta},\mathbf{\Phi})$ and the respective ground-truth maps $(\mathbf{Z}^*_{\text{log}},\mathbf{\Theta}^*,\mathbf{\Phi}^*)$, computed as
\begin{equation} \label{eq:unidepth:orig}
    L_e = \lambda_0 \text{SILog}(\mathbf{Z}_{\text{log}},\mathbf{Z}^*_{\text{log}}) + \lambda_1 \text{MSE}\left((\mathbf{\Theta},\mathbf{\Phi}), (\mathbf{\Theta}^*,\mathbf{\Phi}^*)\right) + \lambda_2 l_{\text{con}},
\end{equation}
where SILog is the standard ``scale-invariant'' loss in logarithmic space used in depth estimation~\cite{eigen2014depth}, MSE is the standard mean-squared-error loss introduced in UniDepth for estimation of dense camera intrinsics, and $l_{\text{con}}$ is the geometric invariance loss utilized in UniDepth for consistency of internal geometric network features to geometric augmentations. The official implementation of UniDepth sets $\lambda_0 = 1$, $\lambda_1 = 0.25$, and $\lambda_2 = 0.1$. Moreover, the initial learning rate is $\alpha = 10^{-4}$. The AdamW optimizer~\cite{loshchilov2017adamw} is employed by UniDepth, with $\beta_1 = 0.9$, $\beta_2 = 0.999$, and weight decay $\lambda = 0.1$.

We recognize that this loss formulation involves one basic empirical loss term, i.e.\ SILog, and two auxiliary terms, i.e.\ MSE and $l_\text{con}$. Furthermore, we observe that normalization of the loss scale to $1$ is not applied, with the effective initial learning rate being $\alpha^{(\text{eff})} = \left(\sum_{i=0}^2 \lambda_i\right) \alpha = 1.35\times{}\alpha = 1.35\times{}10^{-4}$. Thus, throughout our experimentation with \Ours{} on UniDepth, we keep the learning rate at its original effective scale $\alpha^{(\text{eff})}$ to preserve the original loss scale.

\begin{table*}[tb]
    \centering
    \caption{Comparison of \Ours{} vs.\ grid search for optimizing UniDepth~\cite{piccinelli2024unidepth} and its two independent loss HPs for monocular metric depth and 3D estimation. ``Grid'': loss HP values are set in the context of a grid search over two dimensions. Final performance metrics and HPs after convergence are reported for all models.}
    \label{table:unidepth:yoto:vs:grid:only:nuscenes:sunrgbd:mean}
    \resizebox{\linewidth}{!}{%
    \begin{tabular}{lll|ccc|ccc||ccc}
    \toprule
    \multirow{2}{*}{Method} & \multirow{2}{*}{$(\exp(\mu_1),\exp(\mu_2))$} & \multirow{2}{*}{$(\lambda_0,\lambda_1,\lambda_2)$} & \multicolumn{3}{c|}{nuScenes~\cite{nuscenes}} & \multicolumn{3}{c|}{SUN-RGBD~\cite{song2015sunrgbd}} & \multicolumn{3}{c}{Mean} \\
    & & & $\mathrm{\delta}_{1}\uparrow$ & $\mathrm{SI}_{\log}\downarrow$ &  $\mathrm{F}_A\uparrow$ & $\mathrm{\delta}_{1}\uparrow$ & $\mathrm{SI}_{\log}\downarrow$ &  $\mathrm{F}_A\uparrow$ & $\mathrm{\delta}_{1}\uparrow$ & $\mathrm{SI}_{\log}\downarrow$ &  $\mathrm{F}_A\uparrow$ \\
    \toprule
    Grid & $(0.01,0.01)$ & $(0.9804,0.0098,0.0098)$ & 43.8 & 25.32 & 41.1 & 91.5 & 8.24 & 75.7 & 67.7 & 16.78 & 58.4 \\
    Grid & $(0.01,0.1)$  & $(0.9009,0.0090,0.0901)$ & 45.9 & 25.25 & 42.1 & 92.1 & 8.17 & 76.1 & 69.0 & 16.71 & 59.1 \\
    Grid & $(0.01,1)$    & $(0.4975,0.0050,0.4975)$ & 44.8 & 25.86 & 41.5 & \best{92.8} & 8.20 & 76.3 & 68.8 & 17.03 & 58.9 \\
    Grid & $(0.25,0.01)$ & $(0.7937,0.1984,0.0079)$ & 45.5 & 25.23 & 41.1 & 92.0 & 8.12 & 76.1 & 68.8 & 16.68 & 58.6 \\
    Grid & $(0.25,0.1)$~\cite{piccinelli2024unidepth} & $(0.7407,0.1852,0.0741)$ & \scnd{48.5} & \scnd{25.19} & 43.1 & 92.2 & 8.15 & 76.4 & \scnd{70.4} & \scnd{16.67} & 59.8 \\
    Grid & $(0.25,1)$    & $(0.4444,0.1111,0.4444)$ & 45.9 & 25.82 & 42.2 & 92.2 & 8.17 & 76.2 & 69.1 & 17.00 & 59.2 \\
    Grid & $(1,0.01)$    & $(0.4975,0.4975,0.0050)$ & 44.4 & 25.36 & 40.7 & \scnd{92.6} & 8.09 & 76.6 & 68.5 & 16.73 & 58.7 \\
    Grid & $(1,0.1)$     & $(0.4762,0.4762,0.0476)$ & 45.1 & 25.27 & 41.6 & 92.5 & \best{8.08} & \best{76.9} & 68.8 & 16.68 & 59.3 \\
    Grid & $(1,1)$      & $(0.3333,0.3333,0.3333)$ & 48.4 & 25.51 & \scnd{43.4} & 92.3 & 8.17 & 76.5 & \scnd{70.4} & 16.84 & \scnd{60.0} \\
    \midrule
    \Ours{} & $(0.04680,0.04677)$ & $(0.91444,0.04280,0.04276)$ & \best{49.9} & \best{25.07} & \best{43.8} & 92.3 & \best{8.08} & \scnd{76.7} & \best{71.1} & \best{16.58} & \best{60.3} \\
    \bottomrule
    \end{tabular}}
\end{table*}

We compare in Table~\ref{table:unidepth:yoto:vs:grid:only:nuscenes:sunrgbd:mean}: (i) a $3\times{}3$ grid search through normalized, non-learnable loss HPs $(\lambda_0, \lambda_1, \lambda_2) = (\exp(\mu_0), \exp(\mu_1),\exp(\mu_2))/{\sum_{i=0}^2 \exp(\mu_i)}$, where $(\exp(\mu_0), \exp(\mu_1),\exp(\mu_2)) \in \{1\}\times{}\{0.01,0.25,1\}\times{}\{0.01,0.1,1\}$, i.e.\ the search includes the original loss HP combination of UniDepth, and (ii) the \Ours{} optimization of the loss HPs combined with AdamW, analogously to Algorithm~\ref{alg:yoto:sgdw}, setting $\rho = 20$ and $\epsilon = 0.1$ and otherwise following the original optimization settings for regular network parameters which were used in UniDepth~\cite{piccinelli2024unidepth} as mentioned above. Moreover, across all runs, the training set is composed of ScanNet~\cite{dai2017scannet}, Argoverse2~\cite{argoverse2021}, and Waymo~\cite{sun2020waymo}--which were all originally used in~\cite{piccinelli2024unidepth} too---for a total of ca.\ 700K images, and we train the same ViT-L~\cite{dosovitskiy2020vit} backbone as in~\cite{piccinelli2024unidepth} on mini-batches of size 64 for 300K iterations, which takes 20 RTX4090 days. Despite the relatively smaller number of training images and iterations compared to the original 3M and 1M of~\cite{piccinelli2024unidepth} respectively, the performance of the model from the grid search that uses the original HPs on the two zero-shot evaluation datasets we use following~\cite{piccinelli2024unidepth}, i.e.\ the outdoor nuScenes~\cite{nuscenes} with 36,114 images and the indoor SUN-RGBD~\cite{song2015sunrgbd} with 4,396 images, largely matches that of the originally trained model in~\cite{piccinelli2024unidepth}, even exceeding it for the key 3D metric of $\mathrm{F}_A$ on SUN-RGBD.

What is more, the \Ours{} model for UniDepth converges to a set of loss HPs and regular parameters which consistently outperforms the best of the 9 grid-search models across the two benchmarks and the three metrics. This fact is not constrained on the $\mathrm{SI}_{\log}$ metric which is identical with the basic empirical training loss, but it extends into the key depth measure of $\delta_1$ and 3D measure of $\mathrm{F}_A$ which are only employed for testing, implying that our novel approach of optimizing HPs on the training set crucially improves inference performance. Note that we minimally experimented with tuning $\rho$ and $\epsilon$ for \Ours{} in this comparison, trying only $\rho \in \{2,20\}$ and $\epsilon \in \{0.01,0.1\}$, even though our entire optimization algorithm is new and no suitable parameter ranges were previously known. Thus, by only training once, we get a better 3D estimation model than any one found by brute force among an order of magnitude more models, with a significant margin of 0.7\% in $\delta_1$ and 0.5\% in $\mathrm{F}_A$.

\begin{figure*}[tb]
  \centering
  \vspace{-0.35cm}
  \subfloat{\includegraphics[width=0.33\textwidth,trim={0.3cm 14cm 0.5cm 0.3cm},clip]{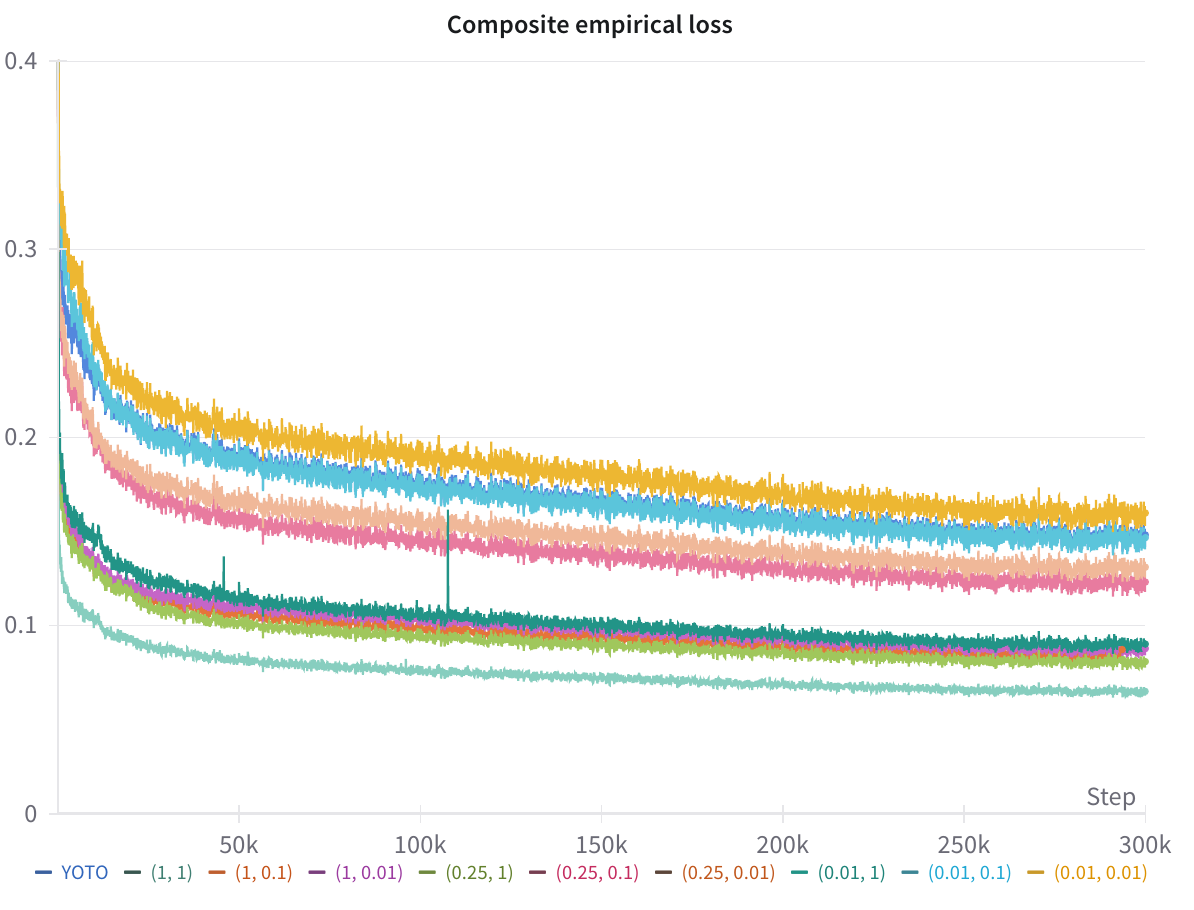}}
  \hfil
  \subfloat{\includegraphics[width=0.33\textwidth,trim={0.3cm 14cm 0.5cm 0.3cm},clip]{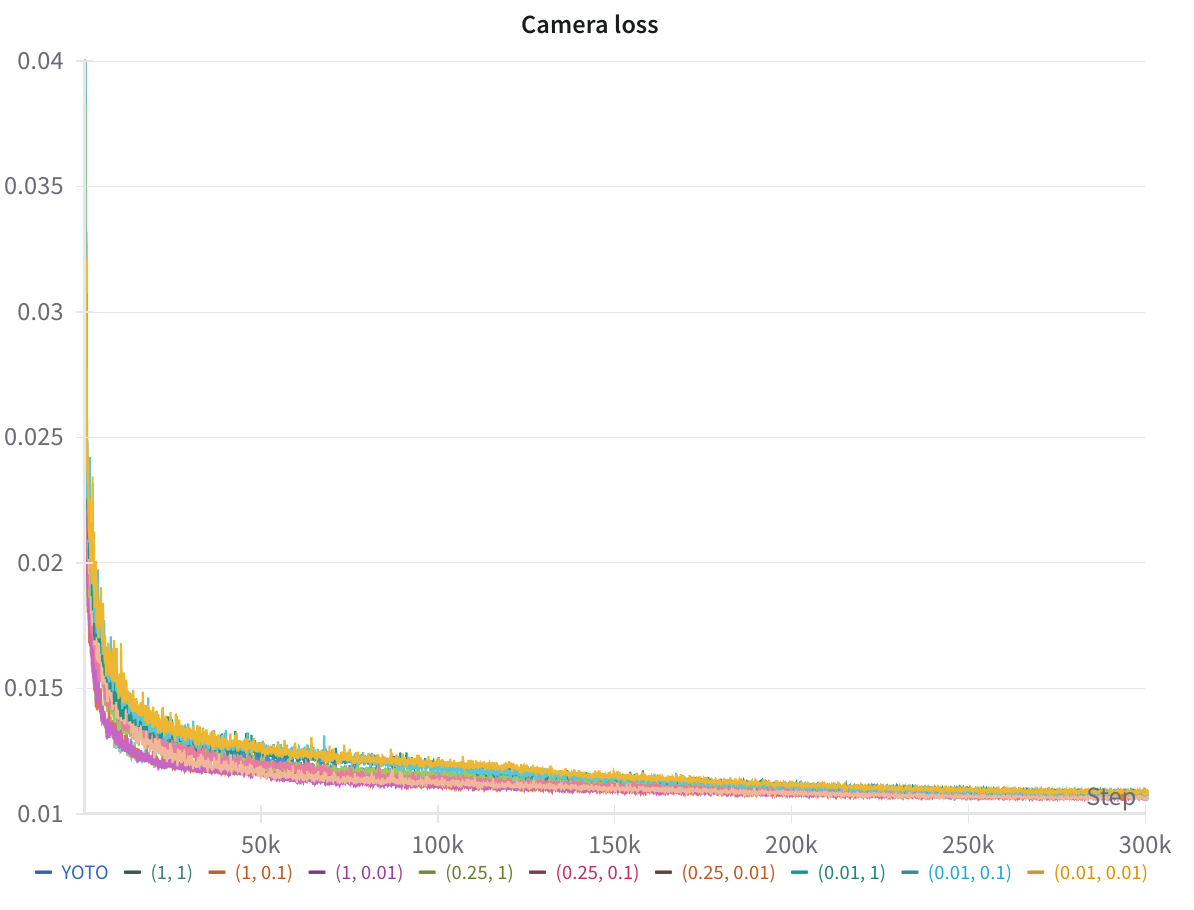}}
  \hfil
  \subfloat{\includegraphics[width=0.33\textwidth,trim={0.3cm 14cm 0.5cm 0.3cm},clip]{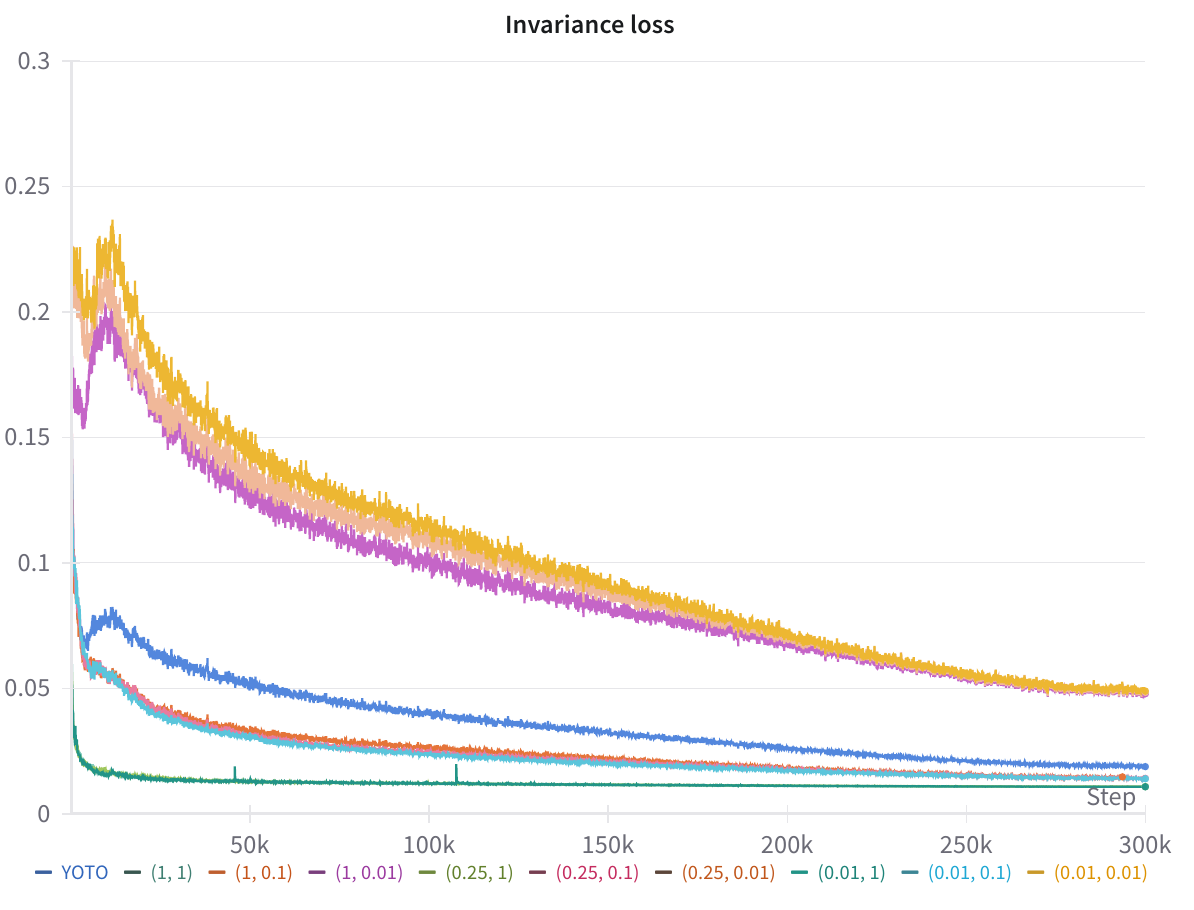}}
  \\
  \vspace{-0.35cm}
  \subfloat{\includegraphics[width=0.33\textwidth,trim={0.3cm 14cm 0.5cm 0.3cm},clip]{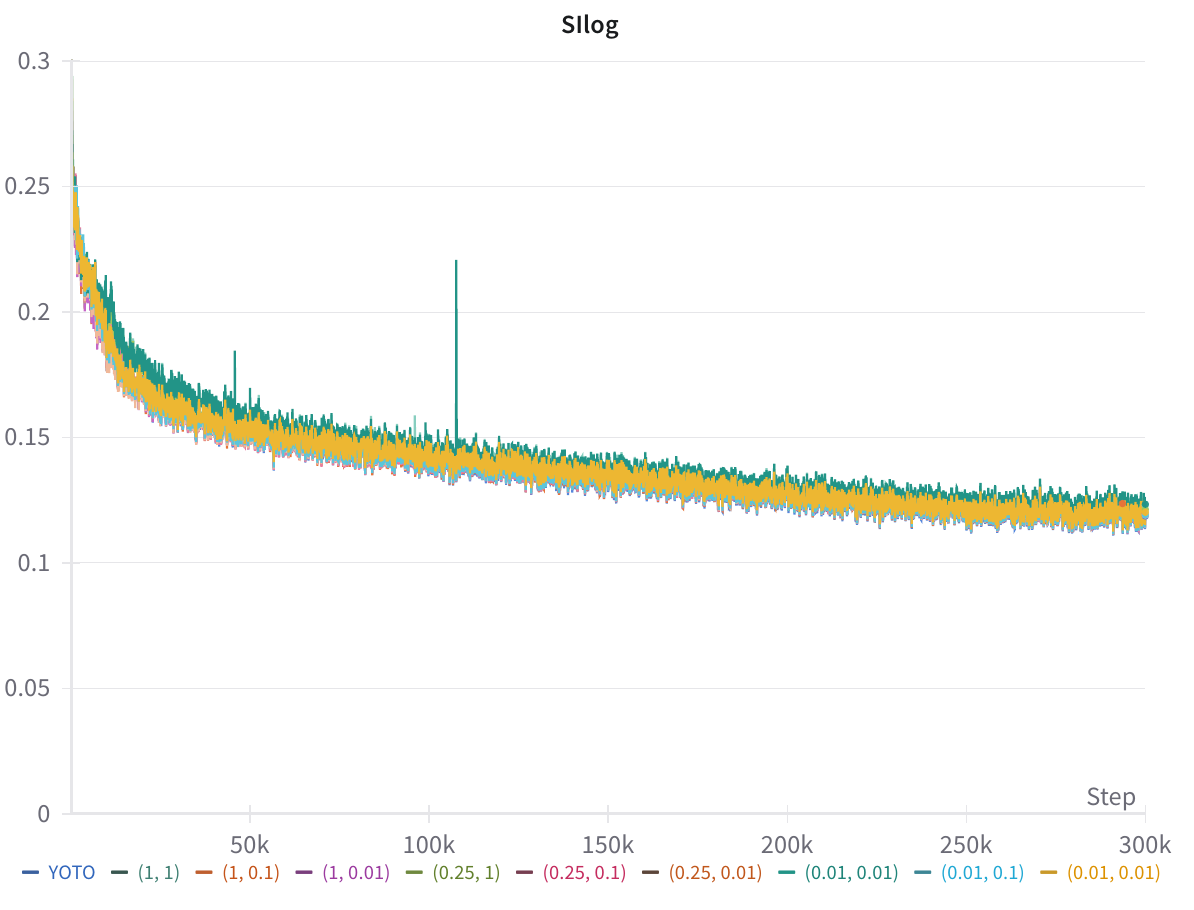}}
  \hfil
  \subfloat{\includegraphics[width=0.33\textwidth,trim={0.3cm 14cm 0.5cm 0.3cm},clip]{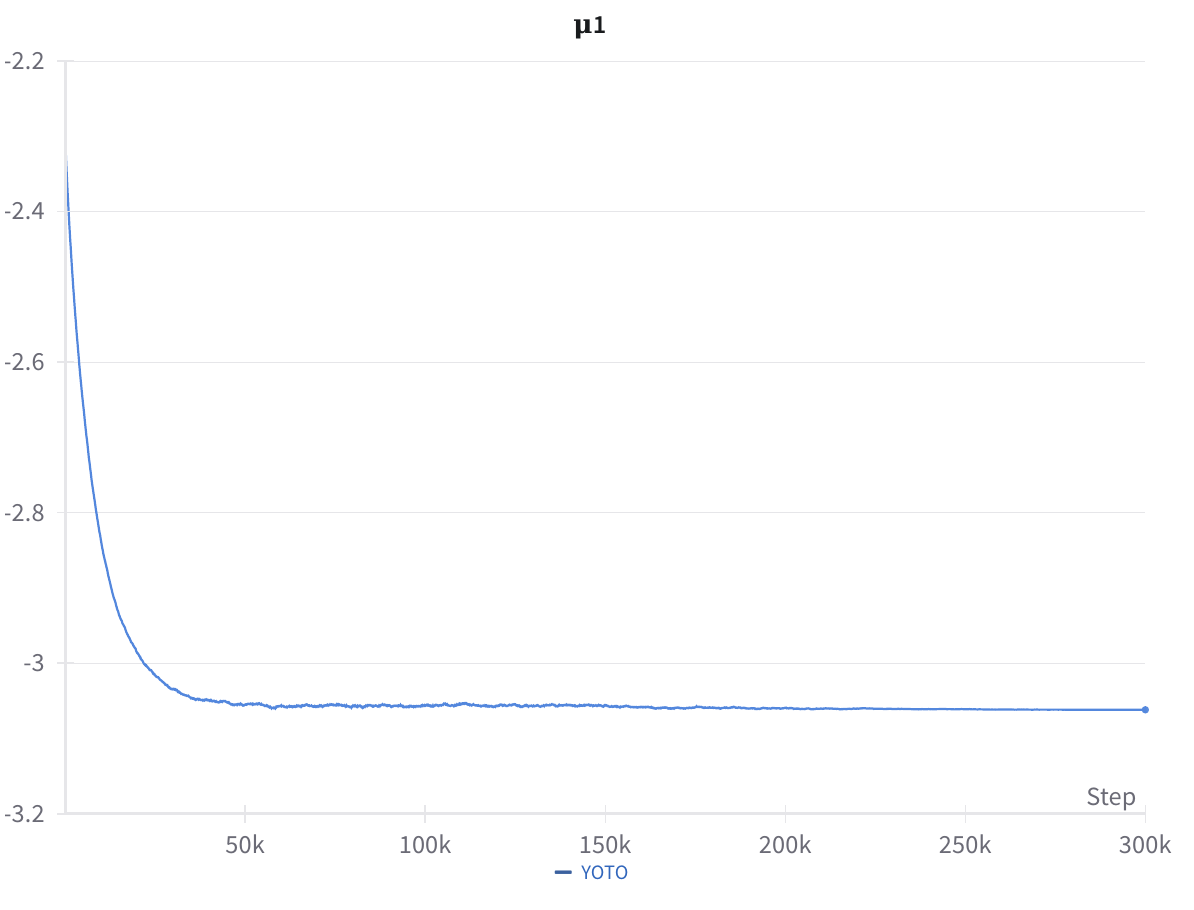}}
  \hfil
  \subfloat{\includegraphics[width=0.33\textwidth,trim={0.3cm 14cm 0.5cm 0.3cm},clip]{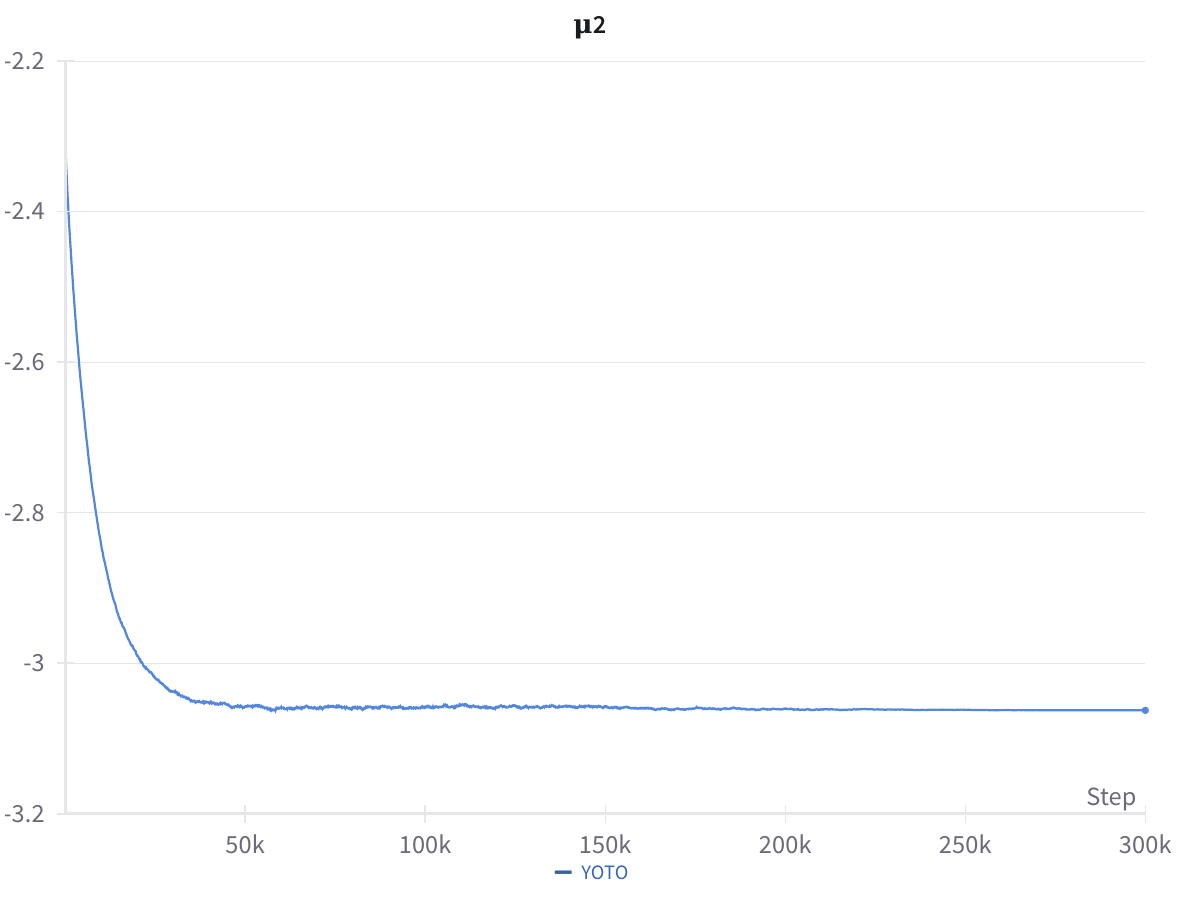}}
  \\
  \vspace{-0.35cm}
  \subfloat{\includegraphics[width=0.33\textwidth,trim={0.3cm 14cm 0.5cm 0.3cm},clip]{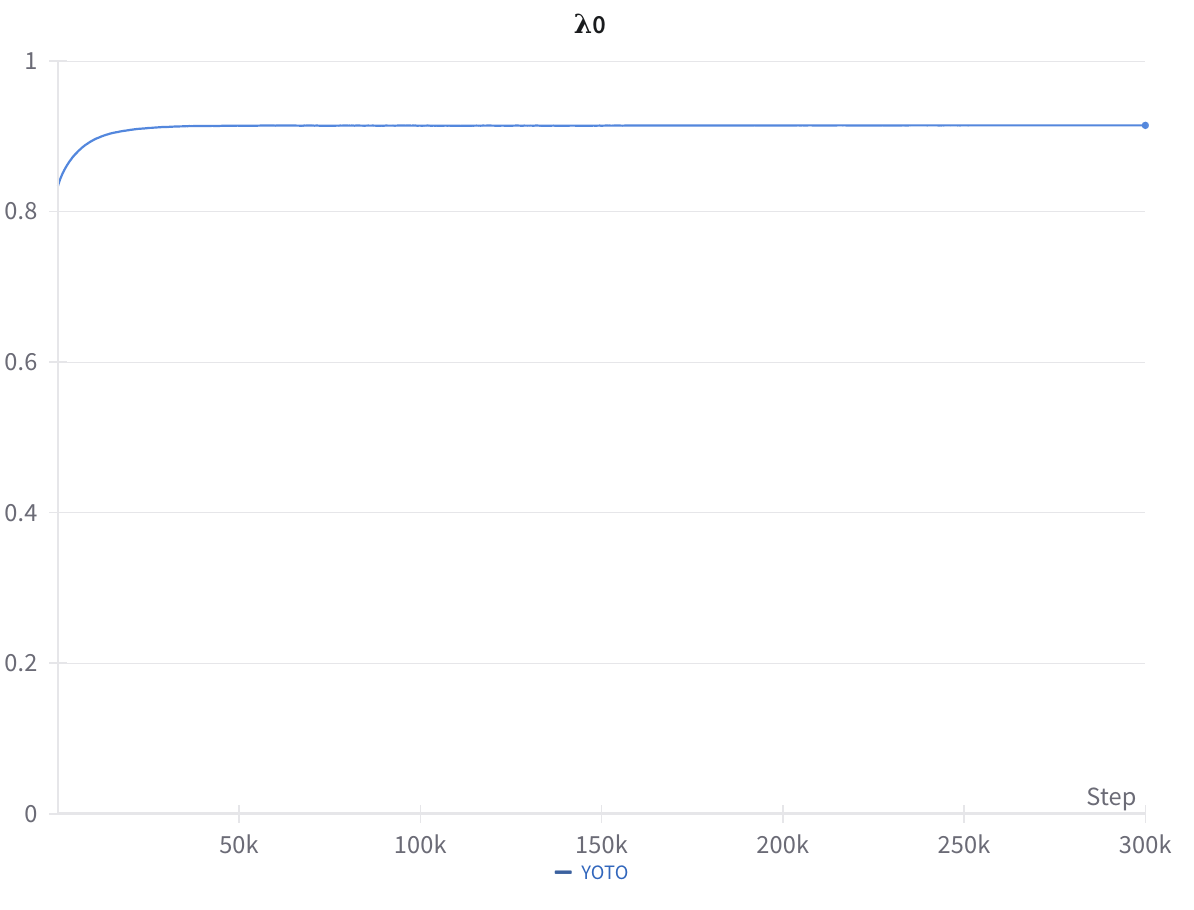}}
  \hfil
  \subfloat{\includegraphics[width=0.33\textwidth,trim={0.3cm 14cm 0.5cm 0.3cm},clip]{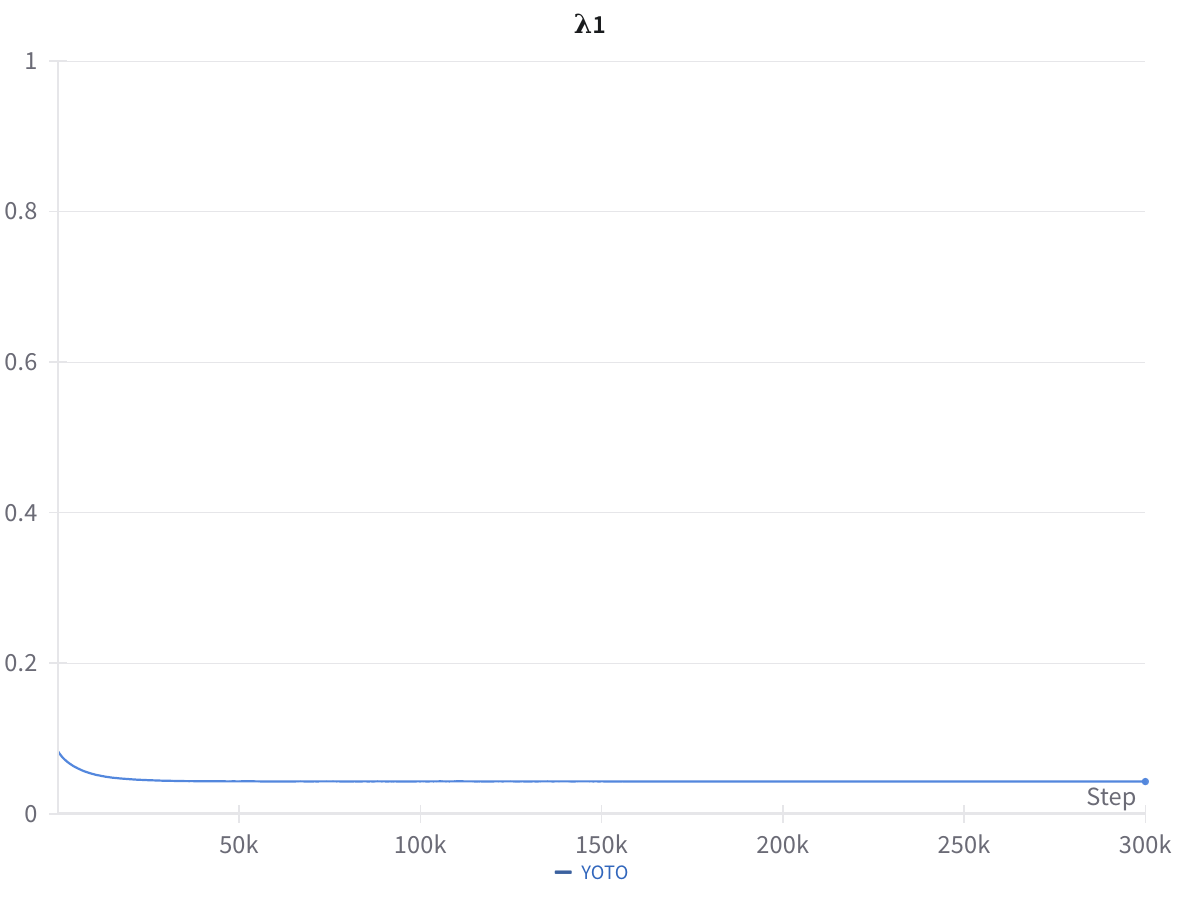}}
  \hfil
  \subfloat{\includegraphics[width=0.33\textwidth,trim={0.3cm 14cm 0.5cm 0.3cm},clip]{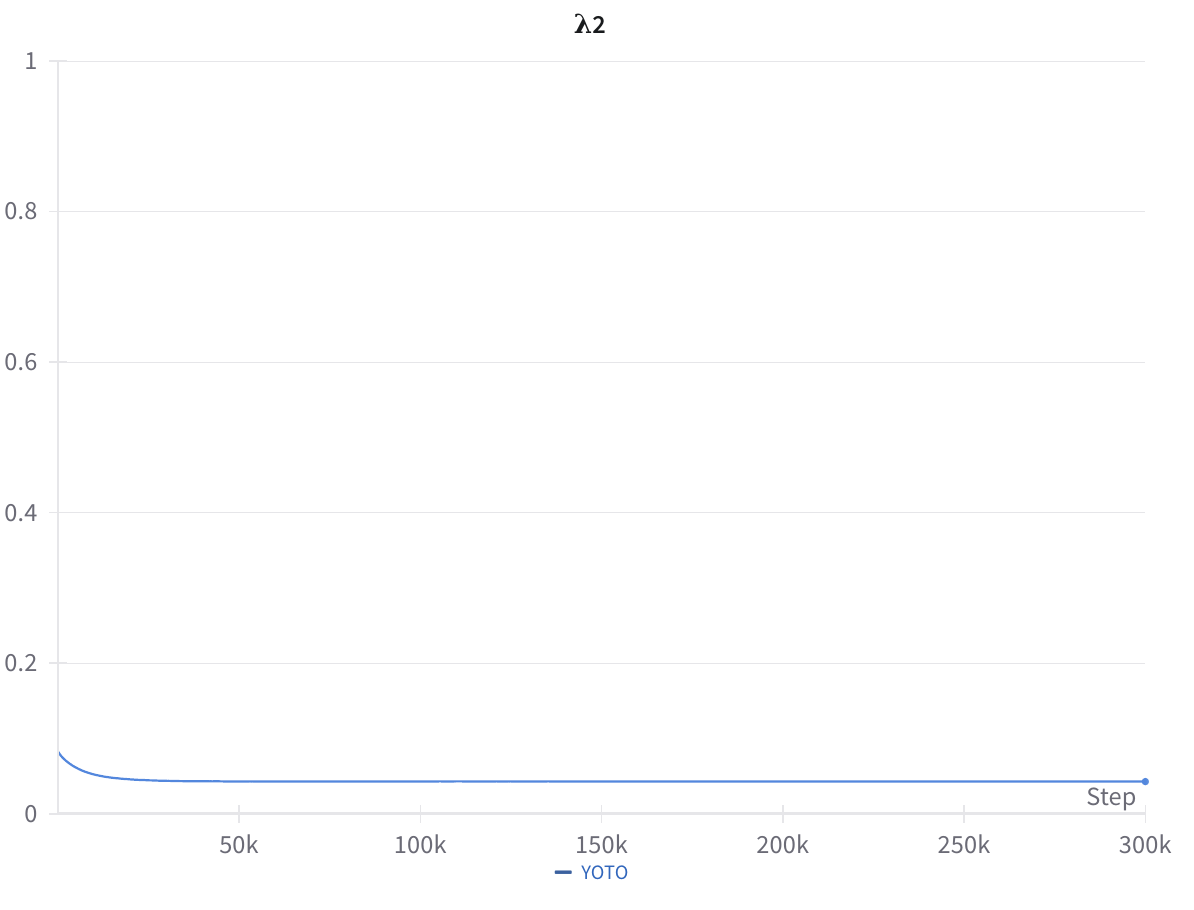}}
  \caption{Comparison of optimization dynamics of \Ours{} vs.\ grid search models for UniDepth~\cite{piccinelli2024unidepth}. Top row: composite empirical loss $L_e$, camera loss $\text{MSE}$, and geometric invariance loss $l_{\text{con}}$, middle row: $\text{SILog}$ loss, $\mu_1$ for \Ours{}, and $\mu_2$ for \Ours{}, bottom row: $\lambda_0$, $\lambda_1$, and $\lambda_2$ for \Ours{}. Grid search models are referred in the legends by their respective constant $(\exp(\mu_1),\exp(\mu_2))$ values. Best viewed on a screen at full zoom.}
  \label{fig:unidepth:yoto:vs:grid}
\end{figure*}

\begin{figure*}[tb]
  \centering
  \vspace{-0.35cm}
  \subfloat{\includegraphics[width=0.33\textwidth,trim={0.3cm 15.5cm 0.5cm 0.3cm},clip]{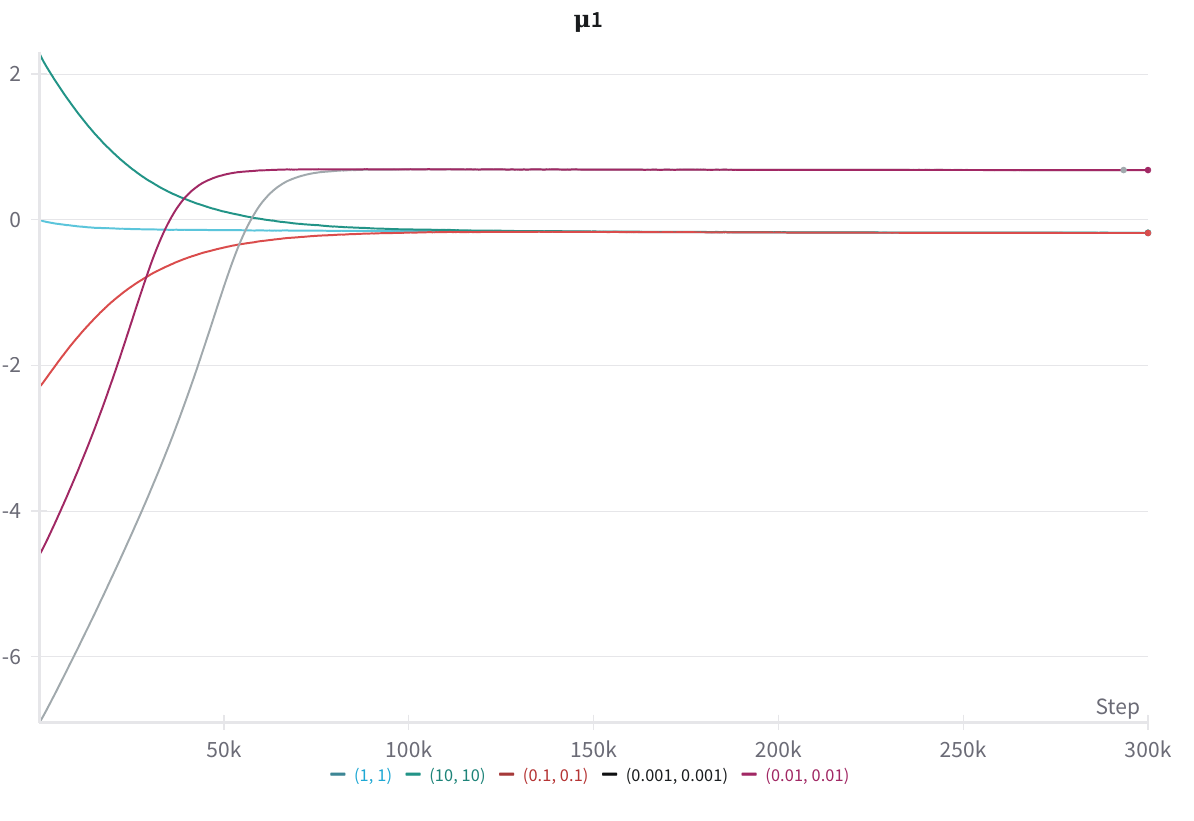}}
  \hfil
  \subfloat{\includegraphics[width=0.33\textwidth,trim={0.3cm 15.5cm 0.5cm 0.3cm},clip]{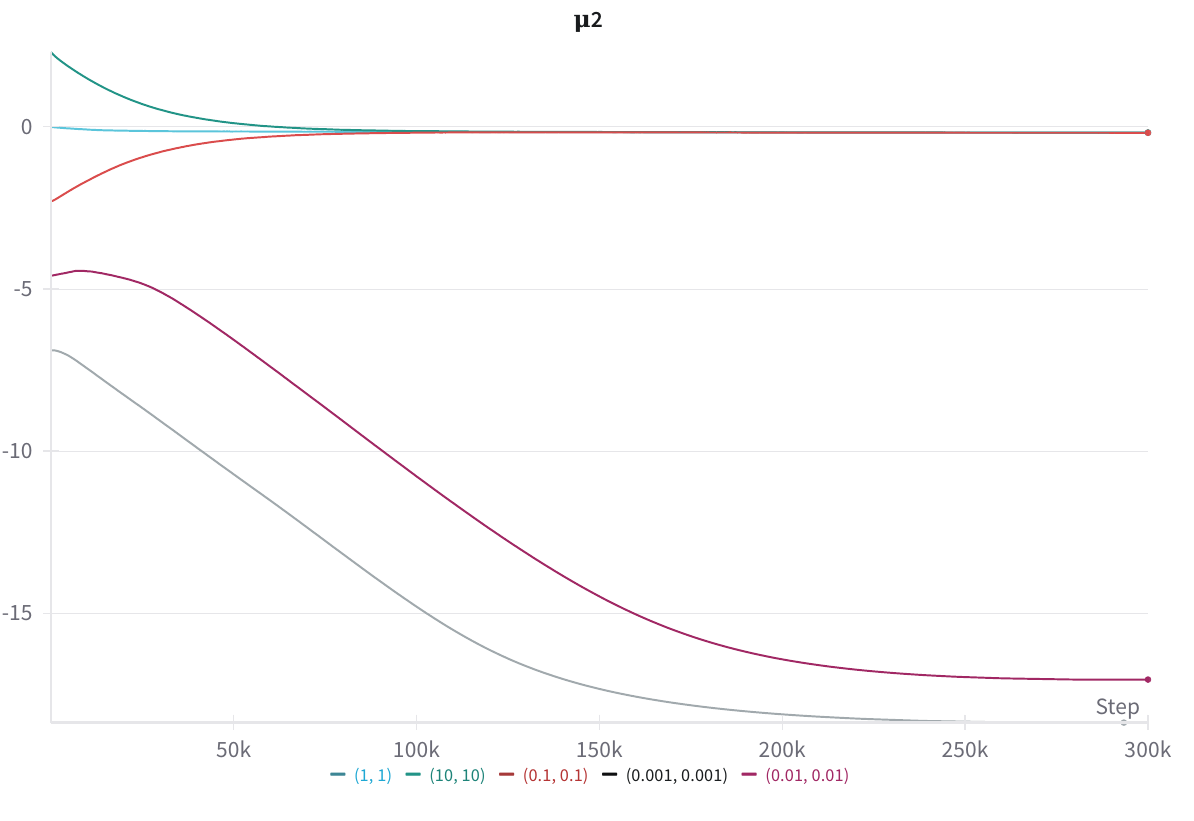}}
  \hfil
  \subfloat{\includegraphics[width=0.33\textwidth,trim={4cm 8.5cm 4cm 9.8cm},clip]{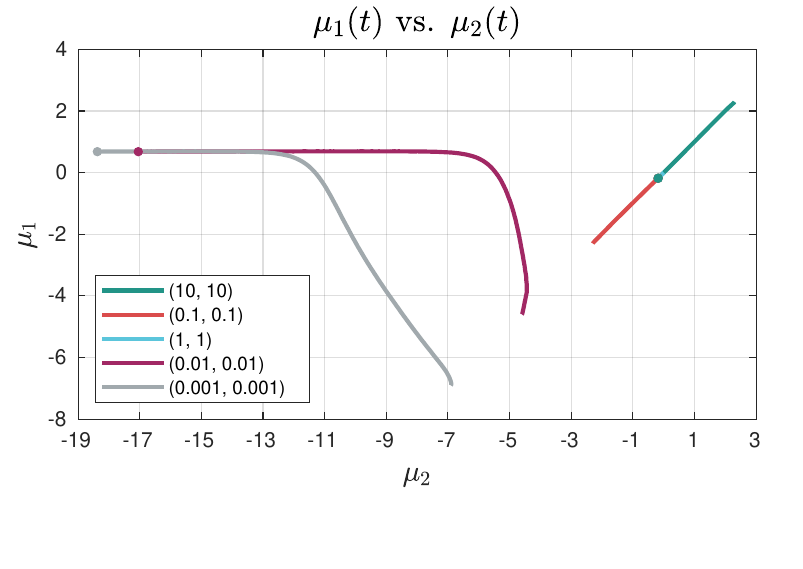}}
  \\
  \vspace{-0.35cm}
  \subfloat{\includegraphics[width=0.33\textwidth,trim={0.3cm 15.5cm 0.5cm 0.3cm},clip]{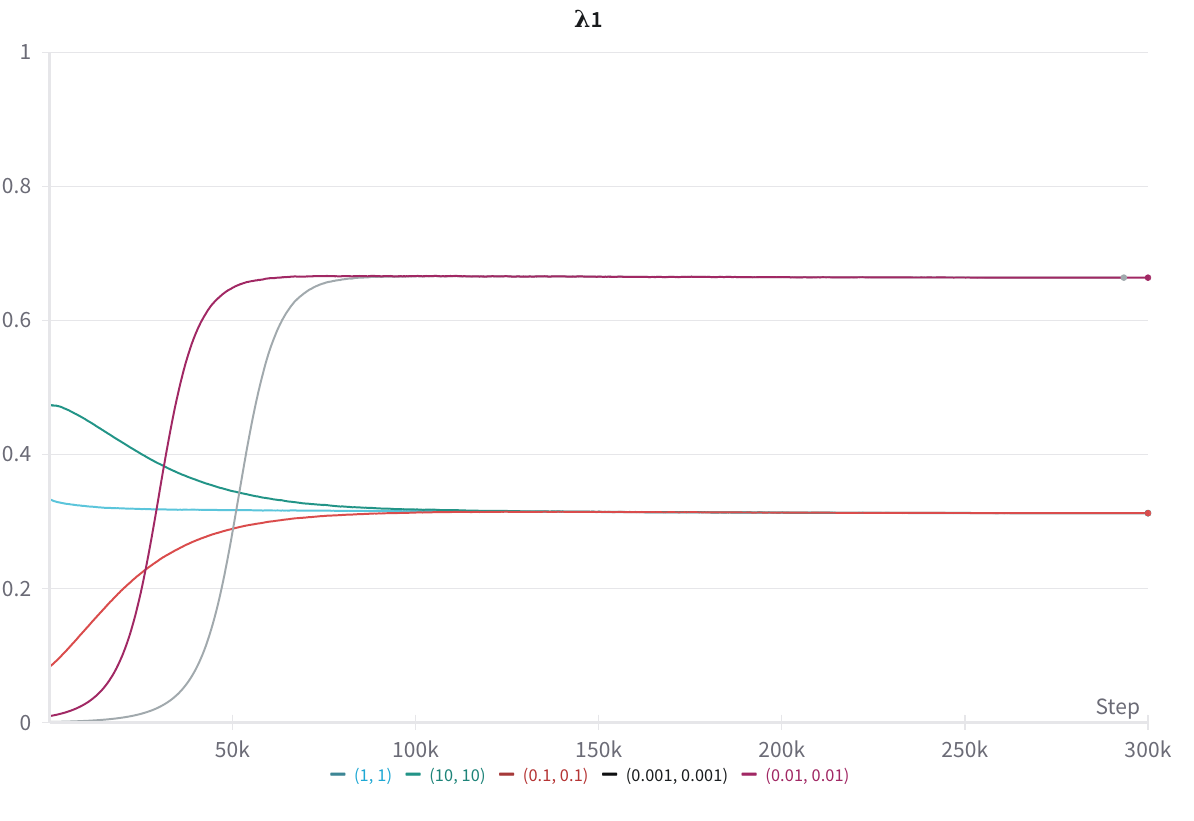}}
  \hfil
  \subfloat{\includegraphics[width=0.33\textwidth,trim={0.3cm 15.5cm 0.5cm 0.3cm},clip]{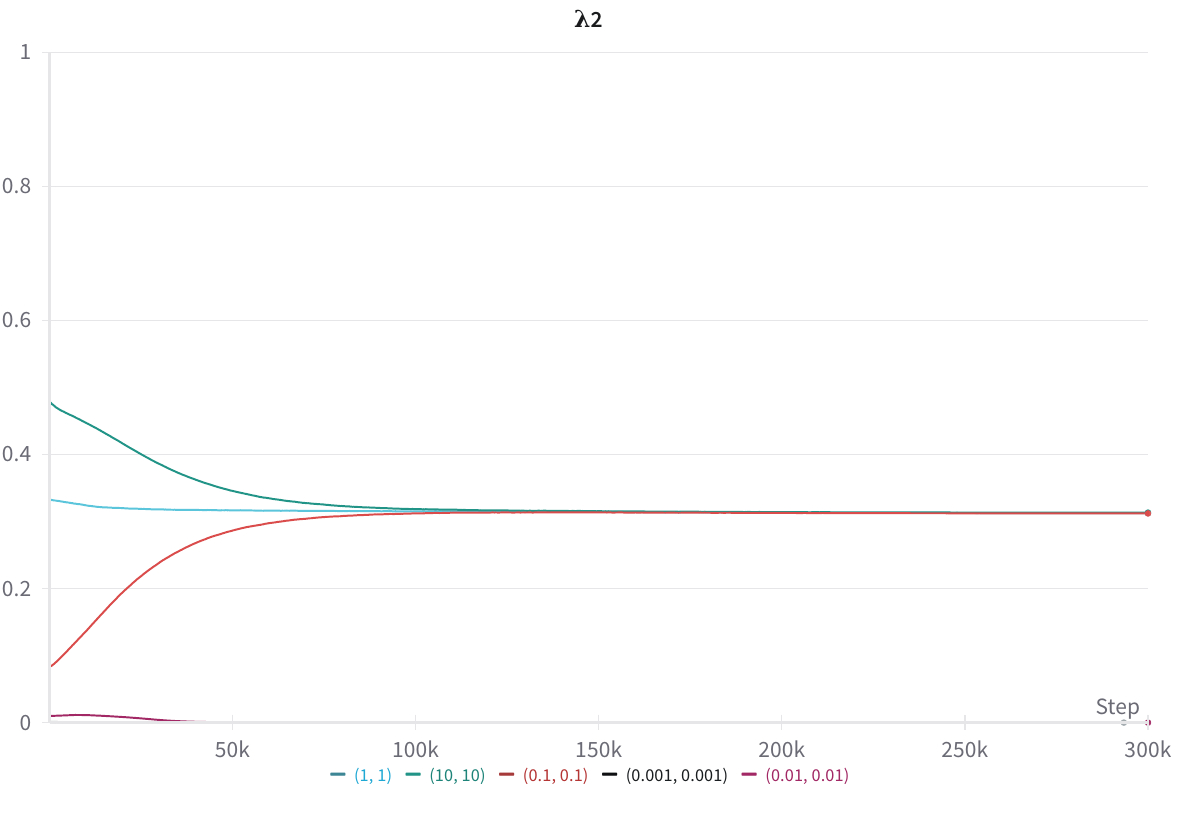}}
  \hfil
  \subfloat{\includegraphics[width=0.33\textwidth,trim={0.3cm 15.5cm 0.5cm 0.3cm},clip]{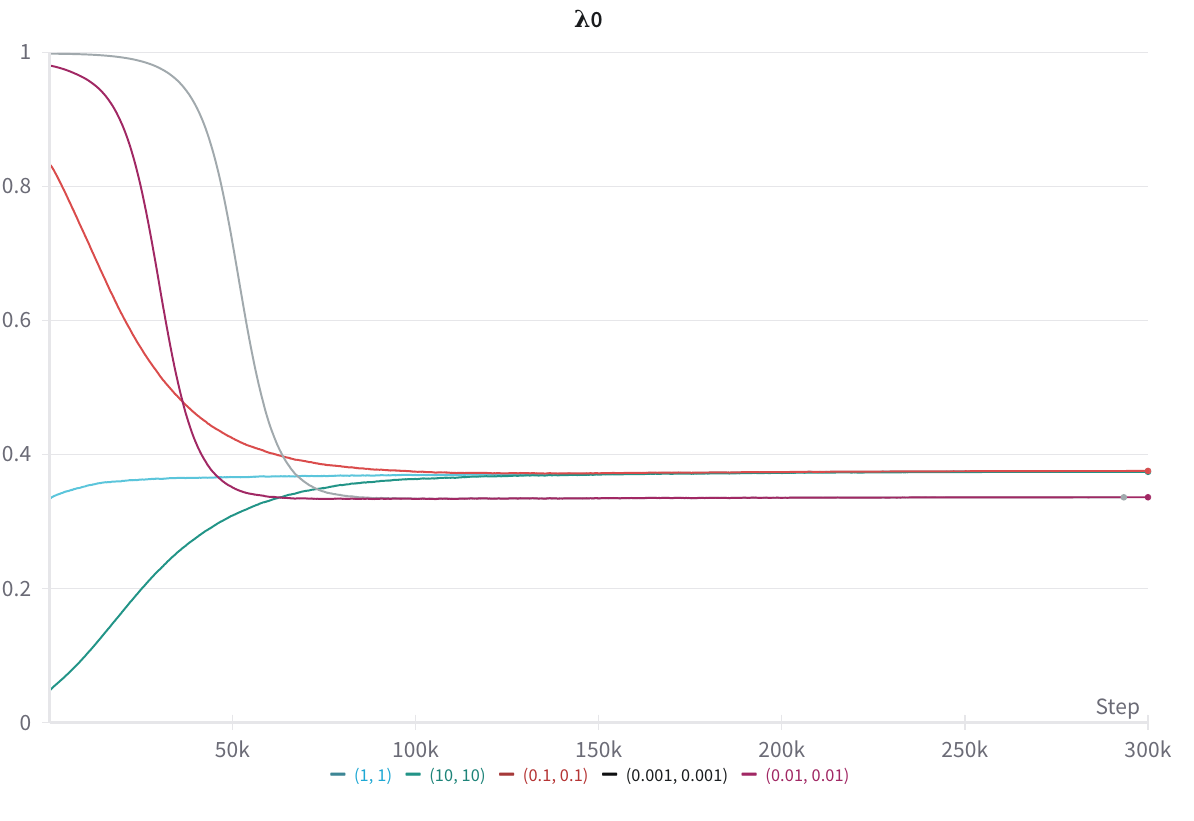}}
  \caption{Analysis of sensitivity of \Ours{} to initialization, performed for training UniDepth~\cite{piccinelli2024unidepth}. We vary the initialization parameter $\epsilon$ for different \Ours{} models while keeping hyperparameter decay fixed at $\rho=2$, and plot the evolution of HPs throughout optimization. Different models are referred in the legends by their respective $(\epsilon,\epsilon)$ values which initialize $(\exp(\mu_1),\exp(\mu_2))$. Top row: $\mu_1(t)$, $\mu_2(t)$, and $\mu_1(t)$ vs.\ $\mu_2(t)$ with final points as dots, bottom row: $\lambda_1(t)$, $\lambda_2(t)$, and $\lambda_0(t)$. Best viewed on a screen at full zoom.}
  \label{fig:unidepth:insensitivity:to:init}
\end{figure*}

We illustrate the detailed optimization dynamics for the above \Ours{} model for UniDepth against those of the compared models from the $3\times{}3$ grid search in Figure~\ref{fig:unidepth:yoto:vs:grid}. We first observe that the \Ours{} model (in blue) exhibits the same dynamics w.r.t.\ the basic empirical loss, i.e.\ $\text{SILog}$, as the grid search models, for which $\lambda_0$ is fixed. Different models exhibit different dynamics in their auxiliary camera and invariance losses, which are generally inversely proportionally high to the corresponding weight $\lambda_i$. \Ours{} automatically discovers a favorable tuple for its adjustable HPs in the initial stage of the optimization and stabilizes at it for the subsequent, longer stage of the optimization.

The importance of initialization of our algorithm motivates us to conduct a sensitivity analysis of this aspect of it in Figure~\ref{fig:unidepth:insensitivity:to:init}. In particular, we train 5 different \Ours{} models for UniDepth, by varying the initialization parameter $\epsilon$ linearly in log space from $10^{-3}$ to $10$ while keeping hyperparameter decay fixed at $\rho = 2$. We observe that for these initializations, each of the 5 models converges to one of two points in the HP space, despite their potentially different initialization, which evidences that \Ours{} is relatively insensitive to initialization. Models with sufficiently high $\epsilon$ converge to a point which is still close to the $\mu_1 = \mu_2$ line, ca.\ at $(-0.18,-0.18)$, with both auxiliary losses contributing non-negligibly at convergence. On the other hand, models with lower $\epsilon$ converge to a point away from the $\mu_1 = \mu_2$ line on which they are initialized, for which $\exp(\mu_2) \approx 0$. In essence, the latter models have automatically selected only the camera loss as a helpful auxiliary loss, and have ``rejected'' the invariance loss, the contribution of which is negligible to them at convergence. Between the two sets of models, we have observed that performance of either is better on one of the two zero-shot evaluation sets and worse on the other, implying that the two convergence points represent different optima both in the HP and in the regular parameter space.

\subsection{Semantic Segmentation}
\label{sec:exp:segmentation}

The second key vision task we examine is semantic segmentation, using the state-of-the-art domain-generalizing method of CISS~\cite{sakaridis2025ciss}. In particular, we use the simpler yet highly performing DeepLabv2~\cite{DeepLab:v2} version of CISS and its basic formulation with losses computed only from the source domain, referred to as CISS-source. In particular, CISS-source originally optimizes a composite empirical loss $L_e$ for an encoder-decoder segmentation network $f = \omega \circ \phi$, where $\phi$ is the encoder and $\omega$ the decoder. Based on (i) a pair of internal feature tensors that the network produces using a pair of images $I_s$ and $I_t$, i.e.\ $\phi(I_s)$ and $\phi(g(I_s,I_t))$ where $g$ is a non-learned mapping, (ii) the softmax output $f(I_s)$ for the first image $I_s$, and (iii) the respective ground truth $Y_s$, we compute
\begin{equation} \label{eq:ciss:source:orig}
    L_e = \lambda_0 l_{\text{CE}}(f(I_s), Y_s) + \lambda_1 l_{\text{inv}}(\phi(I_s),\phi(g(I_s,I_t))),
\end{equation}%
where $l_{\text{CE}}$ is the standard cross-entropy loss which is used in semantic segmentation~\cite{FCNs:segmentation} and $l_{\text{inv}}$ is a feature invariance loss that penalizes differences between corresponding elements of the two internal feature tensors. The original implementation of CISS sets $\lambda_0 = 1$, $\lambda_1 = 10$, and the (constant) learning rate to $\alpha = 2.5\times{}10^{-4}$, and uses the SGDW optimizer~\cite{loshchilov2017adamw}, with $\beta_1 = 0.9$ and weight decay $\lambda = 2$. The formulation of \eqref{eq:ciss:source:orig} includes one basic empirical loss term, i.e.\ $l_{\text{CE}}$, and one auxiliary term, i.e.\ $l_{\text{inv}}$. Again here, the loss scale is originally not normalized to $1$, as $\lambda_0 + \lambda_1 = 11$. In this set

\begin{figure*}[tb]
    \centering
    \noindent
    \pgfplotstableread{
    -2.5     44.08   0.36
    -2     44.23   0.58
    -1.5   44.44   0.57
    -1     44.80   0.62
    -0.5   44.96   0.49
    0      44.98   0.63
    0.5    45.00   0.78
    0.75   44.99   0.92
    1.0    44.55   1.00
    1.25   43.66   1.05
    1.5    43.10   0.59
    1.75   40.32   0.73
    2      34.24   0.32
    }\ioucisssourcegridsearch
    \begin{minipage}{0.4\textwidth}
    \parfillskip=0pt
    of experiments, we first re-train the original CISS-source with its original learning rate $\alpha = 2.5\times{}10^{-4}$ but with normalized loss weights such that the original ratio $\lambda_1/\lambda_0 = 10$ is preserved and additionally $\lambda_0 + \lambda_1 = 1$. We compare in Figure~\ref{fig:ciss:source:miou:val:yoto:vs:grid:search}: (i) a grid search through 13 normalized, non-learnable loss HPs $(\lambda_0,\lambda_1)$, where the uniformity of the 1D grid is in the space of the log ratio of the two HPs, $\log_{10}(\lambda_1/\lambda_0)$, and the grid includes the original $\log_{10}$ ratio of $1$ as mentioned above, and (ii) the \Ours{} optimization of the loss HPs combined with SGDW based on Algorithm~\ref{alg:yoto:sgdw}, setting $\rho=200$ and $\epsilon = \exp(-4)$ and otherwise keeping the original optimization settings of DeepLabv2-based CISS detailed above. We train 3 models for every point of the grid search and 3 models for \Ours{}, using in each case a different random seed for each of the 3 runs. The
    \end{minipage}%
    \hfill
    \begin{minipage}{0.56\textwidth}
    \centering
    \resizebox{\textwidth}{!}{%
    \begin{tikzpicture}
    \tikzstyle{every node}=[font=\small]
    \begin{axis}[
        scale only axis,
        width=\textwidth,
        height=0.6\textwidth,
        xmin=-2.5,
        xmax=2.5,
        ymin=32,
        ymax=50,
        axis x discontinuity=crunch,
        axis y discontinuity=crunch,
        xmajorgrids=true,
        ymajorgrids=true,
        ylabel={Mean IoU (\%)},
        xlabel={Loss hyperparameter log ratio $\log_{10}\left(\lambda_1/\lambda_0\right)$},
        xtick={-2.5,-2,-1.5,-1,-0.5,0,0.5,1,1.5,2,2.5},
        xticklabels = {-$\infty$,-2,-1.5,-1,-0.5,0,0.5,1,1.5,2,2.5},
        ytick = {32,35,40,45,50},
        yticklabels = {0,35,40,45,50},
        x tick label style={font=\scriptsize},
        xtick pos=left,
        y tick label style={font=\scriptsize},
        major tick length=0.5ex,
        legend cell align=left,
        legend pos=south west,
    ]
    
    \addplot[
        color=grid-search,
        mark=*,
        mark size=1.5pt,
        thick,
        error bars/.cd,
        y dir=both,
        y explicit,
    ]
    table[
        x index=0,
        y index=1,
        y error index=2
    ] \ioucisssourcegridsearch;
    
    \addplot[
        color=yoto-negentsoftplus,
        mark=*,
        mark size=1.5pt,
        thick,
        error bars/.cd,
        x dir=both,
        x explicit,
        y dir=both,
        y explicit,
    ]
    coordinates {
        (1.3747,45.38) +- (0.00062,0.34)
    };

    \legend{\emph{Grid search for $\log_{10}\left(\lambda_1/\lambda_0\right)$}, \emph{\Ours{}}}

    \end{axis}
    \end{tikzpicture}}
    \caption{Comparison of \Ours{} vs.\ grid search models for optimizing CISS-source~\cite{sakaridis2025ciss} and its loss HPs for domain-generalizing semantic segmentation. Means and standard deviations (1-$\sigma$) of performance ($y$-axis) are plotted for each examined configuration. The mean and standard deviation of the log ratio of optimized HPs with \Ours{} ($x$-axis) is also plotted, but the latter standard deviation is too small, hence imperceptible in the plot.}
    \label{fig:ciss:source:miou:val:yoto:vs:grid:search}
    \end{minipage}
\vspace{-2mm}
\end{figure*}
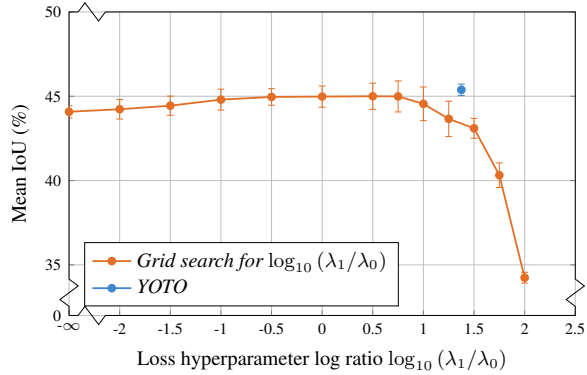

training set comprises the standard domain generalization benchmark of Cityscapes$\to$ACDC~\cite{Cityscapes,ACDC} for normal-to-adverse domain generalization, where only Cityscapes labels are used. We train with mini-batches of size $2$, i.e.\ one image $I_s$ from Cityscapes and one image $I_t$ from ACDC in each mini-batch, for 148,750 iterations, which takes ca.\ 2 RTX4090 days. Validation is performed on the held-out validation set of ACDC using standard mean Intersection over Union (IoU)~\cite{FCNs:segmentation,Cityscapes}. \Ours{} outperforms all models from grid search, even the best one for which $\log_{10}(\lambda_1/\lambda_0) = 0 \Rightarrow \lambda_0 = \lambda_1 = 0.5$. Moreover, \Ours{} exhibits far better robustness to stochasticity, as both its performance variation across runs is much lower than that of grid-search models with fixed HPs and it reliably converges to the same point in HP space with negligible variance, which is at ca.\ $(\lambda_0,\lambda_1) = (0.04,0.96)$. The optimization trajectories of the free HP $\mu_1(t)$ for the 3 runs, shown in Appendix~\ref{sec:appendix:stability}, are also virtually identical. These findings further hint on the stability of our proposed algorithm. To corroborate the generalization improvement in semantic segmentation with one-shot training via \Ours{} against fixed-HP models, we have submitted for online evaluation the predictions on the properly withheld test set of ACDC~\cite{ACDC} of the above best grid-search model using the weights with the median validation performance of the 3 runs as well as the \Ours{} model with the worst validation performance of the 3 runs, i.e.\ favoring the former. Indeed, \Ours{} scores a mIoU of 45.29\% and outperforms the conventional model, which scores 44.52\%, confirming the strength of our joint optimization of regular parameters and HPs in generalization.

\PAR{Discussion and Limitations.} While \Ours{} bypasses the need to manually tune loss HPs, in doing so it introduces two new HPs: (i) the hyperparameter decay $\rho$ which weighs the regularization loss for all $\mu_i$, and (ii) the initialization parameter $\epsilon$. Resp.\ (ii), we argue that most of the existing HPO methods, including model-based, population-based and gradient-based ones, also require some sort of initialization, so this is not a new burden only for \Ours{}, and we have evidenced that our algorithm is fairly robust to the precise choice of $\epsilon$. Resp.\ (i), we argue that $\rho$ is shared across $\boldsymbol{\mu}$, in the same way that weight decay $\lambda$ is shared across $\mathbf{w}$, incurring increasing returns with dimensionality.

\section{Conclusion}
\label{sec:conclusion}

We have presented \Ours{}, an automatic, gradient-based optimization algorithm that tunes loss weight HPs of learned models jointly with regular parameters of the latter in one shot via standard GD. The joint optimization in \Ours{} allows the model to explore a richer parametric space than with existing HPO approaches, leading to consistent gains in final performance at inference besides the obvious gains in efficiency due to the one-shot regime. We view \Ours{} as a first yet firm step in the direction of directly optimizing HPs on empirical losses, which both has wide beneficial implications on the practical experimental optimization of the majority of learned parametric models and opens new avenues for effective and efficient optimization of other types of HPs through standard GD.

\begin{ack}
The author thanks Emmanouil Sakaridis wholeheartedly for the constructive discussions and feedback from him on the formulation of the method and for his valuable pointers to multiple related works. He also thanks Luigi Piccinelli sincerely for his advice and help on reproducing the optimization of the basic UniDepth model.
\end{ack}

{
\small
\bibliographystyle{plainnat}
\bibliography{refs}
}

\newpage

\appendix

\section{Derivation of Composite Empirical Loss Gradients}
\label{sec:appendix:derivation:empirical}

We derive the expression which is provided in \eqref{eq:mu:grad:empirical} for the gradients of our composite empirical loss $L_e$ with respect to the hyperparameters $\mu_i$.

\begin{proof}
Starting from \eqref{eq:softmax}, we apply the quotient rule to get
\begin{multline} \label{eq:proof:mu:grad:empirical:quotient}
    \frac{\partial{}L_e}{\partial\mu_i} =\\
    \frac{\displaystyle\frac{\partial}{\partial\mu_i}  \left(\displaystyle\sum_{j=0}^K \exp(\mu_j)l_j(f(\mathbf{w}))\right) \displaystyle\sum_{j=0}^K \exp(\mu_j) - \left(\displaystyle\sum_{j=0}^K \exp(\mu_j)l_j(f(\mathbf{w}))\right)\frac{\partial}{\partial\mu_i}\left(\displaystyle\sum_{j=0}^K \exp(\mu_j)\right)}{\left(\displaystyle\sum_{j=0}^K \exp(\mu_j)\right)^2}.
\end{multline}
By comparison of \eqref{eq:proof:mu:grad:empirical:quotient} with \eqref{eq:mu:grad:empirical}, we observe that the denominators of the RHSs are identical. Thus, we simply proceed to show the equality of the respective numerators. In particular, we expand the derivatives in the numerator on the RHS of \eqref{eq:proof:mu:grad:empirical:quotient} as
\begin{equation} \label{eq:proof:mu:grad:empirical:differentiation:sums}
    \exp(\mu_i)l_i(f(\mathbf{w}))\left(\displaystyle\sum_{j=0}^K \exp(\mu_j)\right) - \exp(\mu_i)\left(\displaystyle\sum_{j=0}^K \exp(\mu_j)l_j(f(\mathbf{w}))\right).
\end{equation}
Next, we factorize \eqref{eq:proof:mu:grad:empirical:differentiation:sums} as
\begin{equation} \label{eq:proof:mu:grad:empirical:factorization}
    \exp(\mu_i)\left(\displaystyle\sum_{j=0}^K \left(\left(l_i(f(\mathbf{w})) - l_j(f(\mathbf{w}))\right)\exp(\mu_j)\right)\right).
\end{equation}
Finally, the proof is completed by eliminating the zero term in the sum of \eqref{eq:proof:mu:grad:empirical:factorization} for $j=i$ as
\begin{equation} \label{eq:proof:mu:grad:empirical:elimination}
    \exp(\mu_i)\left(\displaystyle\sum_{\substack{j=0 \\ j \neq i}}^K \left(\left(l_i(f(\mathbf{w})) - l_j(f(\mathbf{w}))\right)\exp(\mu_j)\right)\right).
\end{equation}
\end{proof}

\section{Derivation of Regularization Loss Gradients}
\label{sec:appendix:derivation:reg}

We derive the expression which is provided in \eqref{eq:mu:grad:reg} for the gradients of our regularization loss $L_r$ with respect to the hyperparameters $\mu_i$.

\begin{proof}
Starting from \eqref{eq:loss:reg}, we first note that the gradient of the second, softplus term of the RHS is trivial to obtain via the chain rule, so we need to prove that the gradient of the first, negated entropy term on the RHS of \eqref{eq:loss:reg} is equal to the first term on the RHS of \eqref{eq:mu:grad:reg}. We leverage the linearity of the gradient operator to exchange it with the outer sum of the negated entropy term of \eqref{eq:loss:reg}, as well as the product rule over each term of that sum to obtain the following gradient:
\begin{equation} \label{eq:proof:mu:grad:reg:product}
    \displaystyle\sum_{j=0}^K \left(\frac{\partial}{\partial\mu_i}\left(\frac{\exp(\mu_j)}{\displaystyle\sum_{k=0}^K \exp(\mu_k)}\right)\log\left(\frac{\exp(\mu_j)}{\displaystyle\sum_{k=0}^K \exp(\mu_k)}\right) + \frac{\exp(\mu_j)}{\displaystyle\sum_{k=0}^K \exp(\mu_k)} \frac{\partial}{\partial\mu_i}\left(\log\left(\frac{\exp(\mu_j)}{\displaystyle\sum_{k=0}^K \exp(\mu_k)}\right)\right)\right)
\end{equation}
We then break the sum into its $i$-th term and the rest $K$ terms for which $j \neq i$. By calculating away the gradients in \eqref{eq:proof:mu:grad:reg:product} and performing eliminations, the $i$-th term of the sum becomes
\begin{equation} \label{eq:proof:mu:grad:reg:ith:term}
    \frac{\exp(\mu_i)\left(\displaystyle\sum_{k=0}^K \exp(\mu_k) - \exp(\mu_i)\right)}{\left(\displaystyle\sum_{k=0}^K \exp(\mu_k)\right)^2}\left(1 + \log\left(\frac{\exp(\mu_i)}{\sum_{k=0}^K \exp(\mu_k)}\right)\right).
\end{equation}
The rest of the terms of the aforementioned sum from \eqref{eq:proof:mu:grad:reg:product}, after respective calculations of the gradients and eliminations, become
\begin{equation} \label{eq:proof:mu:grad:reg:rest:terms}
    \frac{\displaystyle\sum_{\substack{j=0 \\ j \neq i}}^K \left(-\exp(\mu_i)\exp(\mu_j)\left(1 + \log\left(\frac{\exp(\mu_j)}{\sum_{k=0}^K \exp(\mu_k)}\right)\right)\right)}{\left(\displaystyle\sum_{k=0}^K \exp(\mu_k)\right)^2}.
\end{equation}
We observe that the denominators of both addends in \eqref{eq:proof:mu:grad:reg:ith:term} and \eqref{eq:proof:mu:grad:reg:rest:terms} are identical to that of the first term on the RHS of \eqref{eq:mu:grad:reg}. Thus, we are left to prove that the sum of the numerators of the two aforementioned addends is equal to the numerator of the first term on the RHS of \eqref{eq:mu:grad:reg}. With proper renaming of indices and grouping of sums, these two numerators sum to
\begin{align}
    &\exp(\mu_i)\left(\sum_{j=0}^K \exp(\mu_j)\right) - \sum_{j=0}^K \exp(\mu_i)\exp(\mu_j) -\exp(\mu_i)\sum_{\substack{j=0 \\ j \neq i}}^K \exp(\mu_j)\log\left(\frac{\exp(\mu_j)}{\sum_{k=0}^K \exp(\mu_k)}\right) \nonumber \\
    &{+}\:\exp(\mu_i)\left(\sum_{j=0}^K \exp(\mu_j) -\exp(\mu_i)\right)\log\left(\frac{\exp(\mu_i)}{\sum_{k=0}^K \exp(\mu_k)}\right) \nonumber \\
    &{=}\;\exp(\mu_i)\left(\sum_{j=0}^K \exp(\mu_j)\left(\log\left(\frac{\exp(\mu_i)}{\sum_{k=0}^K \exp(\mu_k)}\right) - \log\left(\frac{\exp(\mu_j)}{\sum_{k=0}^K \exp(\mu_k)}\right)\right)\right) \nonumber \\
    &{=}\;\exp(\mu_i)\left(\sum_{j=0}^K \exp(\mu_j)(\mu_i - \mu_j)\right).
\end{align}
\end{proof}

\section{Stability of Optimization Trajectories of Hyperparameters Against Stochasticity in Training}
\label{sec:appendix:stability}

\begin{figure*}[tb]
  \centering
  \includegraphics[width=\textwidth,trim={0.2cm 25.5cm 1.2cm 0.0cm},clip]{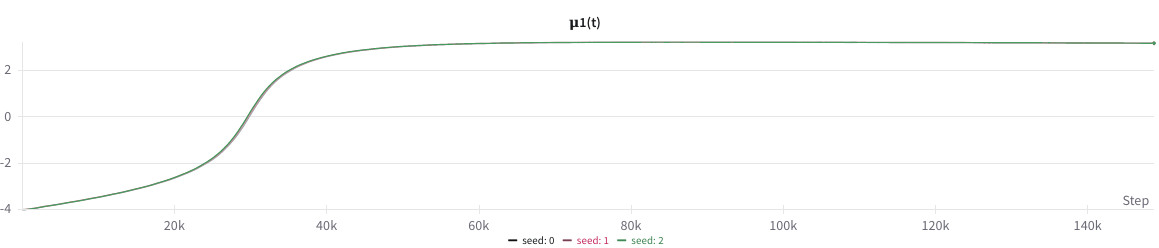}
  \caption{Analysis of sensitivity of \Ours{} to stochasticity in training, performed for training CISS~\cite{sakaridis2025ciss}. The trajectory $\mu_1(t)$ of the hyperparameter $\mu_1$ over training steps $t$ is plotted for 3 different training runs. The 3 training runs are the same as those used to generate the results of Figure~\ref{fig:ciss:source:miou:val:yoto:vs:grid:search}. Across these 3 runs, we vary the random seed in the set $\{0,1,2\}$. Different models are referred in the legend by their respective seed values. All three trajectories are virtually identical and can only be distinguished by zooming in at the interval between 20k and 40k training steps. Best viewed on a screen at full zoom.}
  \label{fig:ciss:stability:trajectories:over:seeds}
\end{figure*}

We further evidence the robustness of \Ours{} to stochasticity, complementing the results of Figure~\ref{fig:ciss:source:miou:val:yoto:vs:grid:search} which pertain to training CISS~\cite{sakaridis2025ciss} with our method. In particular, beyond the latter results showcasing the stable performance and hyperparameter convergence across 3 runs with different random seeds used for training, we show in Figure~\ref{fig:ciss:stability:trajectories:over:seeds} the 3 optimization trajectories of the free hyperparameter, $\mu_1(t)$, for these 3 different runs. All 3 trajectories are virtually identical with each other, with a very minor variation between them in the ``transition'' phase between 20,000 and 40,000 steps of gradient descent, where $\mu_1$ grows faster. This further corroborates the insensitivity of \Ours{} to stochasticity in training, not only with regard to the converged models, but also with regard to the intrinsic training dynamics.

\end{document}